\documentclass{article}

\usepackage{arxiv}

\usepackage[utf8]{inputenc} % allow utf-8 input
\usepackage[T1]{fontenc}    % use 8-bit T1 fonts
\usepackage{hyperref}       % hyperlinks
\usepackage{url}            % simple URL typesetting
\usepackage{booktabs}       % professional-quality tables
\usepackage{amsfonts}       % blackboard math symbols
\usepackage{nicefrac}       % compact symbols for 1/2, etc.
\usepackage{microtype}      % microtypography
\usepackage{lipsum}		% Can be removed after putting your text content
\usepackage{graphicx}
\usepackage[title]{appendix}
\usepackage{multirow}
\usepackage{doi}
\usepackage{longtable}
\usepackage[
backend=biber,
style=alphabetic,
sorting=ynt
]{biblatex}
\title{Scalable and Weakly Supervised Bank Transaction Classification}

\author{{Liam Toran, \hspace{1mm} Cory Van Der Walt, \hspace{1mm} Alan Sammarone, \hspace{1mm} Alex Keller }}
\renewcommand{\headeright}{}
\renewcommand{\undertitle}{}
\renewcommand{\shorttitle}{Scalable and Weakly Supervised Bank Transaction Classification}

\hypersetup{
pdftitle={Scalable and Weakly Supervised Bank Transaction Classification},
pdfsubject={cs.ML}, %% CHANGETHIS
pdfauthor={Liam Toran, Cory Van Der Walt, Alan Sammarone, Alex Keller},
pdfkeywords={Weak Supervision, Bank Transaction Classification},
}

\begin{document}
\maketitle

\begin{abstract}
	This paper aims to categorize bank transactions using weak supervision, natural language processing, and deep neural network techniques. Our approach minimizes the reliance on expensive and difficult-to-obtain manual annotations by leveraging heuristics and domain knowledge to train accurate transaction classifiers. We present an effective and scalable end-to-end data pipeline, including data preprocessing, transaction text embedding, anchoring, label generation, discriminative neural network training, and an overview of the system architecture. We demonstrate the effectiveness of our method by showing it outperforms existing market-leading solutions, achieves accurate categorization, and can be quickly extended to novel and composite use cases. This can in turn unlock many financial applications such as financial health reporting and credit risk assessment.
\end{abstract}

\keywords{
    Weak Supervision \and
    Bank Transaction Classification \and
    Deep Learning \and
    Natural Language Processing \and
    Generative Models \and
    Open Banking Data \and
    Labeling Functions \and
    Anchoring \and
    Time Series \and
    Recurrent Neural Networks \and
    Embeddings \and
    Transfer Learning \and
    Real-time Machine Learning Architecture \and
    Machine learning in Finance \and
    FinTech
}

\section{Introduction}
Accurate bank transaction classification has a wide range of applications in the financial industry. Its derived insights can serve as a pillar for personalized products aimed at consumers, including financial coaching, pre-incident alerting, subscription alerting, reward programs, and personalized credit product matching. On the banking side, traditional credit scores are based only on past credit product usage, which can result in a delayed view of financial health. In contrast, bank transactions can offer precise and granular insights into user spending behavior, bank balances, and risky financial behaviors like overdrafts. Such information could be used to improve credit accessibility for new patrons and small businesses that lack an established credit history, opening up new credit opportunities and enhancing traditional credit scores by providing comprehensible and real-time financial metrics.

However, traditional approaches like manual labeling or rules-based systems are no longer sufficient due to the increasing volume and complexity of transactions. As a result, machine learning algorithms, including deep neural networks, have emerged as promising solutions for categorizing transactions. Such machine-learning-based classifications of transactions face a major challenge: the lack of training labels and the varied and often obscure nature of transaction descriptions. This has prompted the need for scalable approaches that can handle large volumes of unlabeled transactional data while still providing accurate classifications.
In this paper, we present a scalable and weakly supervised transaction classification system utilizing a combination of unsupervised transaction text embeddings, noise-aware label generative models, and deep neural networks. Our novel approach leverages domain knowledge and heuristics to generate broad probabilistic labels that can be used instead of ground truth labels. This allows us to train powerful supervised discriminative models such as deep neural networks for transaction classification, even in the absence of labeled data.

To demonstrate the efficacy of our method, we compare results with the Plaid API, the current leading provider of transactional categorizations. While matching performance for existing accurate categories, our solution can outperform Plaid by an average of 20 points for complex tasks. Although there is still room for improvement in the techniques  used, our work provides a strong foundation for further research and product applications of weak supervision, particularly in the domain of transaction categorization and finance.

\section{Goals and Constraints}
\label{sec:goal_constraints}

Our primary objective is to tackle the complexity of gathering insights from transactional data. In this section, we delineate the problem statement and objectives surrounding our work and the boundaries within which we operate.

\subsection{Transactional Data}

\begin{figure}[ht]
	\centering
	\includegraphics[scale=0.8]{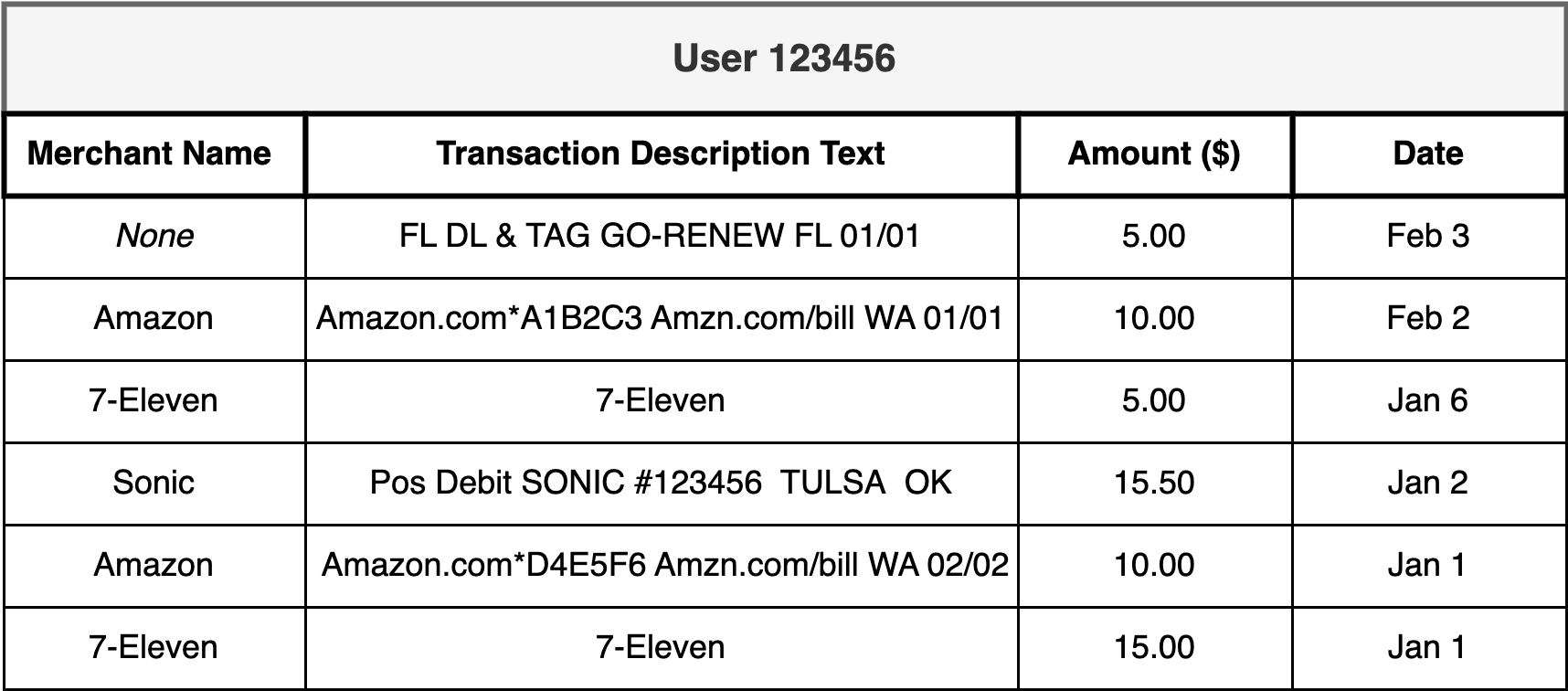}
	\caption{Sample of transaction data.}
	\label{fig:transactions_example}
\end{figure}

Transactional data refers to data associated with common day-to-day banking transactions between patrons and entities such as retail and online stores, financial services such as banking, loans and mortgages, bank transfers, and more. As shown in Figure \ref{fig:transactions_example}, the data includes details such as the date of the transaction, the amount transacted, a text description created by banking systems, and often the other party involved. However, transactional data does not naturally include granular categorizations that help determine which transactions are payments for utilities, rent, mortgages, recurring subscriptions, food and dining, and other such categories. 

Transaction classification presents significant challenges. The sheer volume of transaction data can be overwhelming, as millions of transactions can take place within a single day. Transaction data often contains sensitive information, and privacy concerns may limit the accessibility and usability of this data for annotation. Importantly, the transaction descriptions can be complex and diverse, with transactions varying widely in their characteristics depending on the context. They are often brief, coded, or otherwise obfuscated, making them hard to understand without specific domain knowledge. Manual annotation of such data would be labor-intensive and time-consuming, and automated annotation systems such as the one described in this paper need to be adapted to such volume and complexity. These factors combined make the task of annotating transaction data a daunting one.

Traditional machine learning systems require large sets of labeled data to conduct supervised learning. Small amounts of labels can be obtained, for example by paying annotators or utilizing crowdsourcing and conducting surveys to identify key identifiers for certain types of transactions. However, obtaining these labels is difficult and expensive, and the amount of labels is not sufficient for supervised learning. This is all the more problematic for modern data-hungry methods such as neural networks and highly unbalanced problems such as transaction classification\footnote{Many categories, including those corresponding to monthly transactions such as rent, have a \textasciitilde 1\% positive rate.}. 

This lack of labels naturally calls for techniques that can bypass annotation. Multiple neighboring learning paradigms exist to do so. Firstly, unsupervised learning can be used to learn representations that can be tuned to be relevant to the task at hand \cite{schmarje2021survey}. Semi-supervised approaches use a small number of annotated samples and a larger unlabeled dataset to conduct classifications \cite{olivier2018}. Weak supervision relies on functional heuristics to create a set of annotations for all data points, which are then aggregated using probabilistic models to obtain weak labels that are used to supervise discriminative models \cite{ratner2017data}. Other techniques propose an iterative refinement of annotation sources and have proven successful in applications such as financial fraud detection \cite{zhang2021} and speech recognition \cite{xu2020iterative}.

In this paper, we propose the usage of weak supervision in the context of transaction classification to bypass the main issue of label annotation. Weak supervision allows to generate probabilistic labels from heuristics, rules, or any other guesses and annotations of the data, which will then be used in lieu of manually annotated labels to supervise classifiers. We further showcase how to construct and train a deep neural network transaction classifier utilizing the generated probabilistic weak labels.

Both our classification neural network and weak supervision sources are enhanced using learned unsupervised word representations of our data \cite{joulin2016bag}. We show experimentally that such pre-training techniques can learn powerful associations of the banking system's transactional descriptions. This further confirms the usage of embedding techniques for sequence-based tasks outside of purely natural language, as has been already extensively showcased in ML literature \cite{athiwaratkun2017}\cite{kotios2022}.

We will aim to classify transactions using both spending patterns and transaction text descriptions. Such multimodal approaches are common in financial applications \cite{lv2019}. In the methods and techniques proposed, no annotations are used to fit our models. Only small annotated datasets, consisting of 500 transactions or less, help in the first part to calibrate our different methods, and in the second part to assess their accuracies. This is an acceptable compromise and a very small fraction compared to classical supervised neural network classifiers. To put into perspective, \cite{crichton2017ner} benchmark convolutional neural network techniques on 15 biological datasets each consisting of 4,000 to 84,000 annotated samples. Their study benchmarks performance decrease when the count of annotated training samples is reduced. Specifically, they report an average relative performance decrease of 4\% when the number of training samples is halved, an 11\% decrease when a quarter of the full training set is utilized, and a 22\% decrease when only using a tenth of the annotations.

\subsubsection{Plaid transactional data and Plaid categories}
The transactional data used in this paper originates from the Plaid API. Plaid \cite{Plaid2023Transactions} is a financial technology company that provides a platform for connecting financial institutions with third-party applications. Plaid uses bank-level security protocols to ensure the privacy and security of users' financial data, which are only shared with financial technological products with the user’s consent.

One of the core products of Plaid is its transactional categorization service, which maps each transaction to a set of predefined categories. These categories include common types of transactions such as groceries, utilities, and entertainment, as well as more specific categories such as ride-sharing and streaming services. Plaid transactional categories are designed to be consistent across institutions and provide a standardized way of organizing and analyzing transactional data. In some cases, these categories can be highly precise, such as in the case of telecoms, but in others, such as rent, categorizations contain many errors \hyperref[sec:results]{(4)}. These noticeable errors motivated this paper’s work of improving transaction classification.

\subsubsection{Validation and testing annotation sources}
A small number of annotations are used in the context of this work only to calibrate and assess the performance of our models. Even fairly deterministic transactions such as telecom bills can have outliers that make accurate labeling challenging. In the case of isolated transactions such as manual payments, classifying transactions from text descriptions alone is intractable, and only transacted amounts and payment frequencies can give clues for labeling. As a result, the only authority that can label transactions with full accuracy is the original user who made the payment.

To annotate our validation and testing sets, transactions were randomly sampled from our dataset for the purpose of manual human annotation. Clues were obtained from parsing text descriptions and spending patterns. In some instances, the descriptions were ambiguous and clues alone could not ascertain the sample’s labels with certainty, in which case the sample was not annotated.

For more ambiguous and unbalanced categories such as rent, positive labels were obtained through a customer survey. The customers identified key characteristics such as payment method, day of month payable, transaction amount and contracting body, and in some cases even transaction description. This information could then be matched against their transactional record to identify positive samples. Some curation was included such as bounding spending amounts.

Due to these many constraints surrounding labeling, the end-result validation and testing annotations used in this paper are imperfect, both from a size and quality perspective. Manual effort was put into curing mislabels, using multiple annotators, and reducing obvious biases, but gold-standard annotated sets are impossible to guarantee in this use case. 

\subsection{System Objectives}

{Deploying machine learning solutions is a great challenge due to the
many factors and requirements involved. }

\subparagraph{End-to-end design}

{From a collaboration perspective, many actors such as software
engineers, data engineers, product owners, data scientists, machine
learning engineers, operations team, and more, need to be involved. To
add to the exciting nature of cross-functional collaboration, all
of our chosen solutions need to fulfill production constraints.
Amongst these are the
limited time and computing resources available and the need for highly
reproducible pipelines. Error handling, logging, and versioning are also
crucial for real-time user-facing products.
Additionally, operation costs and efficiency need to be taken into
account. Real-time systems provide many unique
challenges, more so in our case given the size of the models and the
volume of data that needs to be processed.}

{Overspecialization leads to more complex and rigid code. General purpose architectures are often simpler and faster to design, not to mention easier to reason about and evolve \cite{ousterhout2018}. There is of course a cost in terms of development time, and one way to balance these needs is to make the interfaces at least somewhat general-purpose. The interface and high-level construction of our system should therefore be model-agnostic and should abstract away from the details that are particular to any one model or task.}

{Achieving these goals is a key part of any machine learning product,
with a well-known fact that most machine learning models do not make it
into production. Thankfully, many tools exist to help practitioners. The
majority of these tools are open-source and built for the machine
learning community, and have helped thousands to deploy their ideas in
the real world. In our use case, we found success using Git for code
version control, DVC for annotated datasets version control and science
experimentation backbone, MLFlow for model artifacts and
metadata storage, Kubeflow Pipelines for orchestration, and Apache Kafka
for data storage in the context of real-time inferences. On top
of these general tools, most of this work uses the Python ecosystem, and
the libraries used in our methodology are mentioned in their relevant
sections.}

\subparagraph{Scaling to new use cases}

{Traditional classification systems often require the redevelopment of
models when incorporating new classification tasks due to their sensitivity and the
difficulty in leveraging previous knowledge when tackling new challenges.
This lack of scalability results in increased development
time, higher costs, and a less accurate system, as new tasks require
significant manual intervention and may negatively impact the
performance of existing models. Transactional data can be divided into a
wide array of categories of interest. The financial industry needs a
scalable and flexible transaction classification system that can easily
integrate new tasks while maintaining or improving overall performance.}

{Use-case scaling motivated our decision to use one binary
classifier per task instead of one multi-task classifier.
Combining many low-prevalence classes into one
multi-task classifier can be especially challenging. The addition
of a new task can impact the performance of the other tasks and classes are difficult to tune in isolation.
Additionally, transaction categories possess a hierarchical structure, with some being mutually
exclusive, and others having overlap or inheritance properties. These
challenges can be solved, however, binary classifiers are much simpler to
implement in practice. When using multiple one-vs-all binary
classifiers, each can be frozen upon reaching satisfying results, and
subsequent work can be focused on adding new use cases without damaging
the performance of the frozen categories.}

\section{System Overview}

{Our transaction categorization system consists of five main layers:
data ingestion, data preparation, label model training and inference,
discriminative model training and inference, and data publishing. Firstly, 
we present the transaction categorization model which encompasses
data preparation, label model, and discriminative model. We then demonstrate how the
categorization model is implemented in a complete system that can ingest
transaction information from various sources, handle training and
inference of the model, and then publish predictions to any downstream
applications.}

  \begin{figure}[ht]
	\centering
	\includegraphics[scale=1]{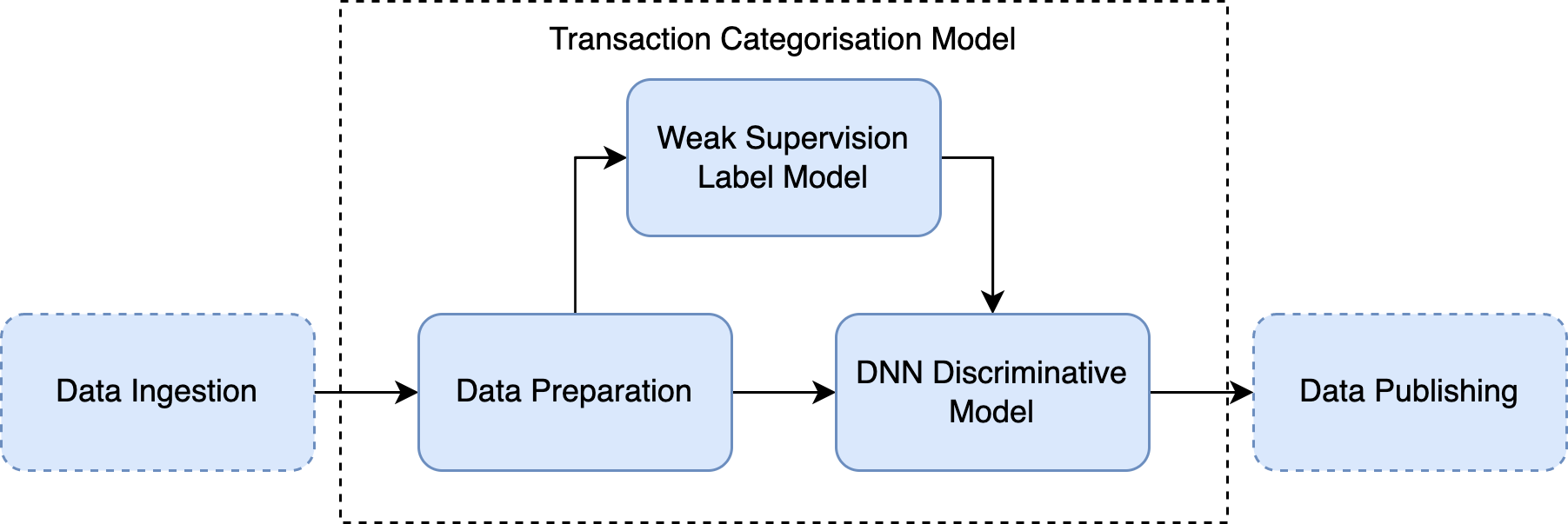}
	\caption{High-level system overview.}
	\label{fig:overview}
\end{figure}

\subsection{Model Pipeline}

{The objective of the model pipeline is to categorize raw transactional data
to create insights that will be used in the context of
financial applications. Transactions first need to be cleaned and
grouped together in order to extract information. Using a set of
constructed heuristics and the grouped transactions, the label model
generates the first set of predicted categorizations, which will serve as
labels for our neural-network-based transaction classifier.}

\subsubsection{Data Preparation}

{Raw transaction descriptions are incredibly noisy and often lack
meaning when considered in isolation. To prepare these raw
transactions for our downstream machine learning approaches, data
preparation is essential and consists of several steps, including data
joining, text preprocessing, and feature engineering.}

\paragraph{Text normalization}

{Text normalization is an essential step of most natural language
processing problems as it allows to smooth out raw text data and reduce
its noise.}

{In our use case, removing this noise has vast benefits, such as
removing personally identifiable data and allowing us to group
transactions with the same text description, which will be incredibly
useful in the following sections, both to reduce the size of our dataset
and utilize aggregate information to classify transactions. This goal of
denoising and aggregating transactions directed the design of our normalization
process to create outputs that are invariant to different types of
noise.}

\begin{figure}[ht]
	\centering
	\includegraphics[scale=0.23]{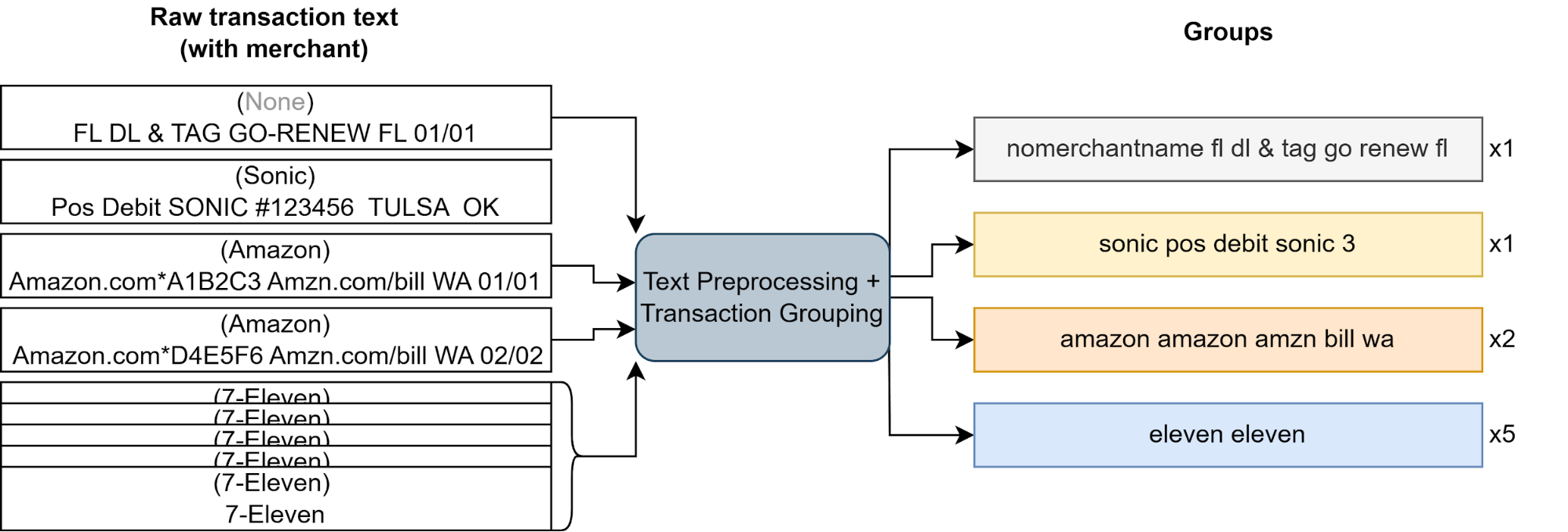}
	\caption{Examples of Normalization and Grouping of
Similar Transactions.}
	\label{fig:normalizations_example}
\end{figure}

{As shown in Figure \ref{fig:normalizations_example}, multiple cleaning steps such as
appending merchant names to the text and smoothing noisy text using
regular expressions are used to normalize the noisy transaction text
input data.}

\paragraph{Data Aggregation}

{The data aggregation stage involves grouping transactions by customer
account and cleaned transaction text. Each group comprises multiple
transactions with identical text for the same bank account. Every
transaction within a group has a specific date and dollar amount
(positive for debits, negative for credits).}

{These groups are constructed under the assumption that two transactions
with the same text for the same user account have the same downstream
category, thereby providing an invariance property to our categorization
problem. This assumption is valid for the vast majority of our data and
allows us to reduce the amount of data while leveraging group-wise time
series information in subsequent pipeline steps. }

{From these groups, we extract the time series of transaction amounts
and apply various aggregations to obtain the maximum, minimum, count,
mean, standard deviation, median, and coefficient of variation of the
groups' transaction amounts. We also compute more comprehensive
statistics on the series' time patterns, such as
the mean time between two consecutive transactions in days.
These aggregates will be used to create heuristics that serve as the input of the
label-generating model}

\begin{figure}[ht]
	\centering
	\includegraphics[scale=0.2]{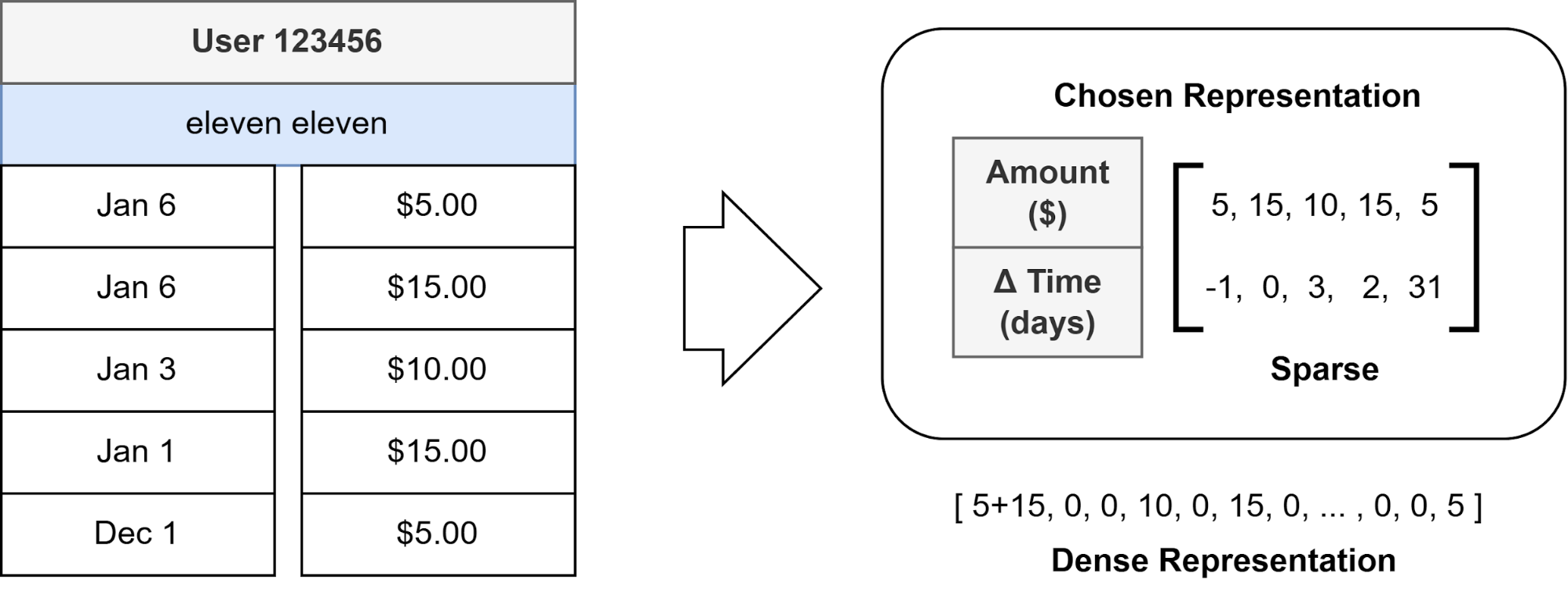}
	\caption{Sparse vs. Dense Transaction Group
Representation}
	\label{fig:sparse_vs_dense}
\end{figure}

{In the downstream discriminative neural network model, these groups need to be
inputted as a sequence. As shown in Figure \ref{fig:sparse_vs_dense}, instead of the
classical approach of creating a fixed-frequency time series, which we
refer to as a ``dense format'', we structure the transaction events $T$ in
the two-dimensional format $T = (amount, \Delta t)_i$, starting from the most
recent transaction (0-th) in the group. $\Delta t_i$
denotes the number of days between the i-th transaction and the (i-1)-th
transaction. This sparse adapted representation allows to efficiently capture
any pattern in the time component, such as multiple transactions on the
same day or extended periods between transactions. Given our usage of
recurrent neural networks, the sparse format also helps with long-term
memory loss and gradient vanishing issues.}

\subsubsection{Transaction Language Processing}
\label{sec:fasttext}

{Groups of transactions share the same text, which for most categories
is the richest feature. Being able to classify this transaction text
frames part of our pipeline in the context of natural language
processing (NLP).}

{Transaction sentences initially consist of characters and words. These texts
need to be converted into numerical representations that can be used
as input features of machine learning models such as neural networks.
Embedding is a technique used to map large vocabularies into lower-dimensional numerical spaces while capturing the underlying semantic
relationships between words or phrases. This mapping ensures that words
with similar meanings have similar vector representations, facilitating
better performance in NLP tasks such as text classification, sentiment
analysis, and machine translation.}

{During development, we experimented with multiple
ways of embedding our transaction descriptions. The baseline for
embedding consisted of a weakly supervised sparse embedding.
This baseline was first improved using pre-trained
(off-the-shelf) English corpus embeddings that are available
online \cite{torchtext}. While the performance of our models was increased by
using these off-the-shelf embeddings, they could not generalize well to
transaction text. }

{Transaction text stems from the English language for the most part but
is also different in many ways. While an English-speaking person can
read most English transaction text, these pre-trained embeddings were not able to
produce the best results due to their training data consisting of only
pure English sentences. To give a concrete example, tokens like
``chkng'' (checking) or ``vrznwrlss'' (Verizon Wireless) are relatively easy for a
human annotator to understand in the context of transaction
classification, but very difficult for available embeddings that were
trained from English language sources such as literature, Wikipedia, or
even the general internet.}

  \begin{figure}[ht]
	\centering
	\includegraphics[scale=0.15]{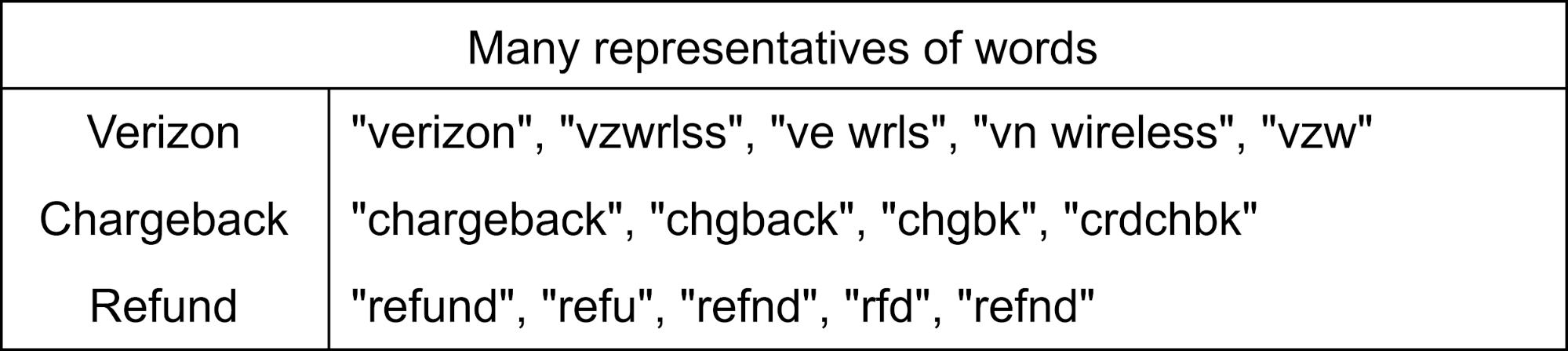}
	\caption{Examples of tokens in the transaction text
describing the same root word}
	\label{fig:root_tokens}
\end{figure}

{We experimented with two methods to learn representations for these
varied transaction tokens. Firstly, by adding a character level weakly
supervised CNN to off-the-shelf embeddings as described in \cite{salur2020}, and secondly,
by retraining unsupervised embedding methods such as FastText \cite{joulin2016bag} on our
transaction corpus. The transaction-trained FastText embedding yielded
large improvements over both the off-the-shelf English corpus embeddings
and the CNN-enhanced embeddings. An in-depth comparison of these
methods can be found in \hyperref[app:ablation]{Appendix \ref{app:ablation}}.}

{FastText \cite{joulin2016bag} is a word embedding method that 
upon techniques like Word2Vec \cite{mikolov2013efficient} 
and can generalize to out-of-vocabulary words by
representing each word as the sum of its n-grams. For instance, the word
``mathology'', which is not part of English vocabulary, would not be
given a meaningful word vector by Word2Vec. FastText however can
represent it as the sum of ``math'' and ''ology'', giving downstream
machine learning models an adequate representation of the token.
Word2Vec is usually preferred amongst practitioners for well-structured
and clean text, e.g. English literature, with FastText used in tasks
with diverse languages, rare words, and misspellings, e.g. text
messages.}

  \begin{figure}[ht]
	\centering
	\includegraphics[scale=0.2]{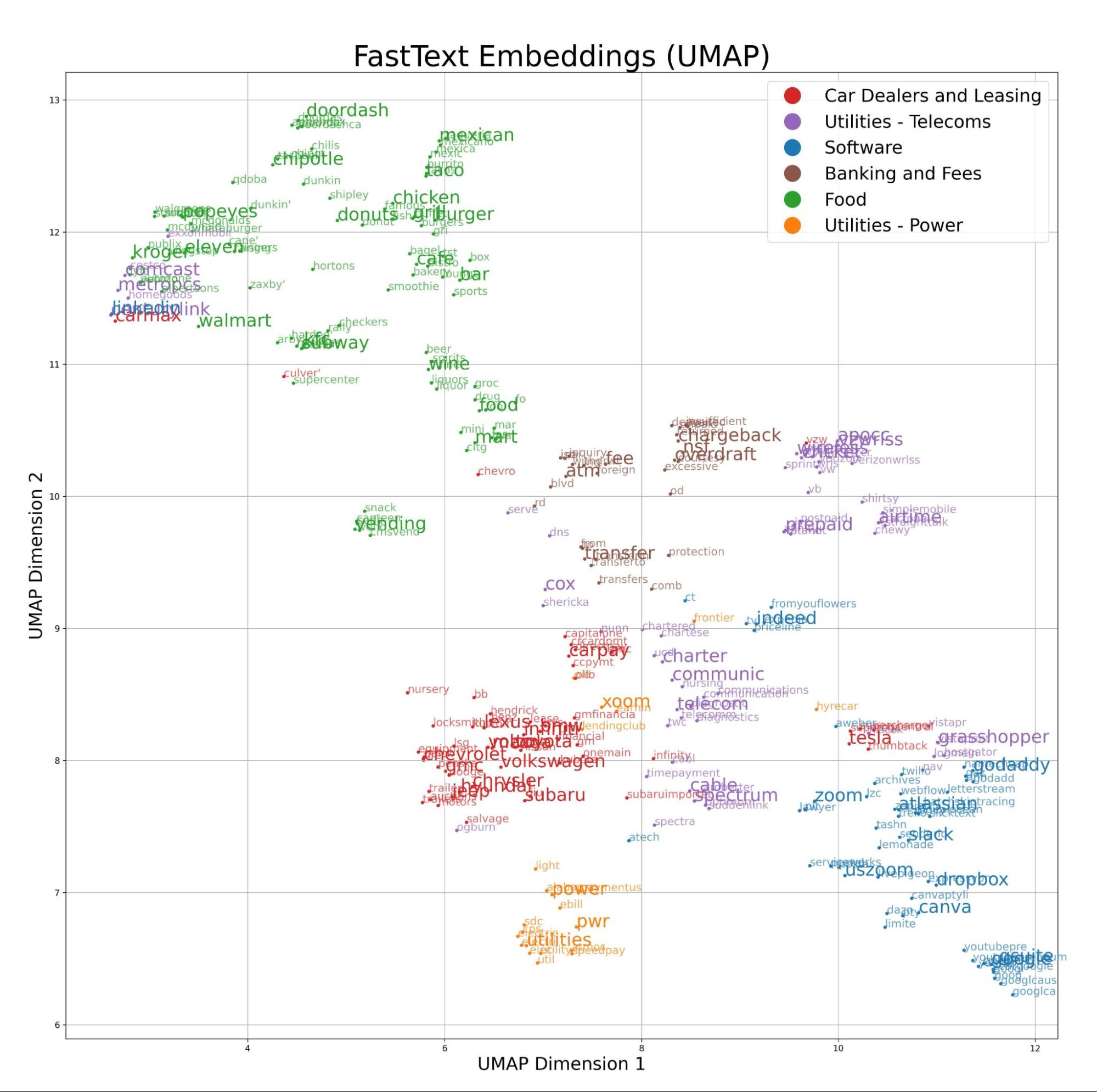}
	\caption{Transaction words and corresponding nearest
neighbors, expanded upon in Appendix \ref{app:fasttext}}
	\label{fig:fasttext}
\end{figure}

{The results of our transaction-trained FastText model were particularly
revealing. The FastText model was able to discern useful associations of
words, effectively clustering words that belong to the same categories
together. This ability to group transaction terms was a pivotal factor
in the overall performance of the text modality of our system, with
results helping both our label model and discriminative model. This
impressive result underscores the efficacy of the FastText method and
the benefits of training a customized unsupervised word embedding for
tackling difficult text tasks such as transaction classification.}

\subsubsection{Weak Label Generation}
\label{sec:weak}

{A weak label refers to an approximate or noisy label associated with a
data point. Unlike ground truth labels, weak labels can be generated
from various sources, such as heuristics, crowd-sourcing, or other
auxiliary information. These weak supervision sources, also named
labeling functions, can be inaccurate, can abstain on portions of the
data, and can agree or disagree with each other, but by combining them
using statistical models, it is possible to create a broader and
more accurate composite weak label.}

{The higher quality, model-generated weak labels, can then be used on
their own to answer the classification task, but are more commonly used
to supervise a higher-capacity machine learning model. Adding
supervised (discriminative) models trained using generated labels often
leads to increased accuracy due to the better classification powers of
the chosen machine learning models, which in essence improve the
decision boundaries of the label-generating (label) model using the
input data. }

{Weak supervision is particularly useful in domains where labeled data
is scarce or difficult to obtain. It has demonstrated success in many
domains such as Natural Language Processing, Computer Vision,
Biomedicine, Pharmaceuticals, and Finance among others.}

\paragraph{Labeling Functions}

{In our transaction classification use case, labeling functions commonly
use either information in the group's normalized transaction text or
its associated spending pattern such as transaction amounts and spending frequency.}

{Ideally, labeling functions should have both high accuracy and high
coverage. However, this is naturally a trade-off. As shown in
\cite{ratner2018training}, special care needs to be given to creating accurate
labeling functions as sources uncorrelated to the task greatly worsen
the convergence of the chosen label model. }

{Multiple ways exist to assess labeling function performance. Firstly,
checking overlaps and coverage of labeling functions can help notice
errors. For instance, when designing a classifier for automobile
payments, the pattern "ford" was used in pattern matching, but triggered
false positives due to the word "afford", which was fixed along with
other examples using word breaks. Another simple way of assessing
accuracy is using an annotated development set. However, this can
sometimes be a luxury as annotations are expensive and descriptions can
be widely varied. For many classes, small annotated sets cannot come
close to representing the task's diversity. Lastly, the label model
learns the conditional accuracies of each labeling function. These learned
accuracies, while not exact, can also be used to detect when the
information given by a labeling function is no better than random. Using
these strategies, labeling functions can be refined or pruned to lead to
better label model performance.}

{Two main modalities of labeling functions are constructed in our use
case. The first type uses the transaction time series and annotates
samples based on spending frequency patterns or lack thereof, for both
dates and amounts. For instance,  these can identify groups of
transactions with a frequency of roughly 15 or 31 days. The second modality
for labeling functions is text. Text-based labeling functions
utilize pattern search to find specific known flags in the transaction text
description, or anchoring which is explained in the next section. }

\paragraph{Anchoring and Fuzzy Word Search}
\label{sec:anchoring}

{Text is typically the strongest signal for transaction categorization.
Therefore, particular attention is dedicated to developing weak label
sources that focus on text features. While transaction labeling
functions based on spending patterns can naturally encompass large
portions of the data, text-based labeling functions, which
traditionally rely on pattern matching, often yield sparse annotations.
For instance, identifying restaurant text flags would require mapping 
upwards of 660,000 \cite{statista2022} existing restaurants names in the
USA.}

{To address the sparse coverage of text-based labeling functions, we
shift our attention to the continuous representations offered by text embeddings.
Techniques such as those used by \cite{chen2022} demonstrated
improvements to generated labels by extending assigned labels to their
nearest neighbors in embedding spaces. In our context, rather than the
resource-intensive process of identifying the nearest neighbors of each
data point,  we utilize the similarity between transaction texts and a
list of word vectors called "anchors".}

{These anchors are selected manually and are representatives of the
classification we aim to achieve, as either positive or negative
exemplars. Each anchor is assigned a similarity threshold that
transaction texts must surpass to be labeled by the corresponding
labeling functions. These thresholds are manually tuned and selected, as
most embedding methods can supply a list of words most similar to a
given anchor.}

{In order to apply word-based similarity to input sentences, we use the
maximum similarity between words in the sequence and the anchor. This
approach is more interpretable and adjustable compared to other pooling
techniques such as averaging the sentence's word vectors. }

{Labeling clusters of data using anchor similarity greatly improved
coverage of our text-based weak supervision sources and the accuracy of our
label and discriminative models. These improvements further show that
our transaction-corpus-trained embedding presented in \ref{sec:fasttext} 
learns similarities that are correlated to our
classification task.}

\begin{figure}[ht!]
	\centering
	\includegraphics[scale=0.275]{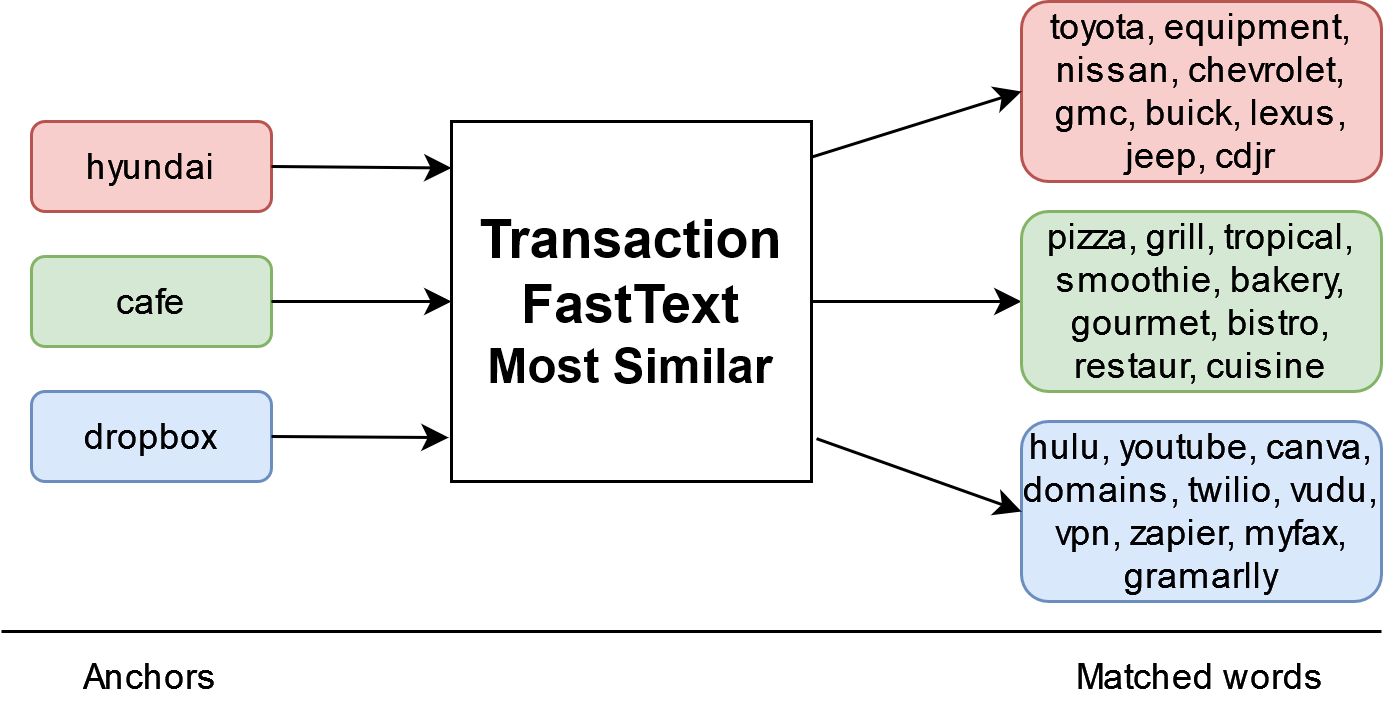}
	\caption{Anchors and corresponding labeling function matched
words.}
	\label{fig:anchors}
\end{figure}

{As shown in Figure \ref{fig:anchors}, this fuzzy word matching utilizing
anchors helps scale each keyword in word search lists to hundreds or
thousands of examples that surpass the chosen similarity thresholds.
This significantly reduces the need to develop exhaustive and expensive
lists of matching patterns for each category we aim to predict,
increasing the performance and scalability of our weakly supervised
approach.}

\paragraph{Label Generating Model}

{Labeling function outputs are used as input for the label-generating
(label) model. The label model combines weak supervision sources to
create probabilistic weak labels that will supervise our downstream
discriminative model.}

{The Snorkel Library\cite{ratnersnorkel} was used to train our label
model. The learning objective of the label model is to estimate the
label-conditional accuracies of each labeling function. It is trained
through gradient descent and uses a loss function that is based on the
reconstruction error between the estimated accuracies' inferred overlaps
and coverage and the provided training weak label sources overlaps and
coverage. These estimated accuracies are combined with a given class
balance prior to obtain the output joint probabilities of the observed
labeling function annotations and the true labels.}

{While the Snorkel library documentation refers to MeTaL \cite{ratner2018training}, 
a method that uses a graphical model known as a factor graph to
represent the dependencies between weak supervision sources and the true
labels to generate the labels, this process is currently not fully
implemented in the open-source Snorkel library and its factor graph
defaults to the trivial assumption of Naive Bayes. Moreover, abstaining
labeling functions do not update the predicted probabilities of the
label model.}

{By generating joint probabilities, the label model can 
handle a wide variety of weak supervision sources, including noisy,
incomplete, and conflicting labels, as well as rules and heuristics that
may not be entirely reliable. These probabilities are used to supervise
the following deep neural network.}

\subsubsection{Discriminative Deep Neural Network}
\label{sec:dnn}

{The label model's predicted probabilities are only a step above using
heuristics and often don't generalize well to complex decision
boundaries or input noise. In order to better utilize our transaction
text and time series data, we chose a deep neural network (DNN)
architecture for our discriminative model.}

\paragraph{DNN Architecture}

{It has been shown \cite{ratner2018training}\cite{radford2022}
that DNNs trained using label-model generated labels can approach
the performance of traditionally supervised DNNs using large hand-labeled
datasets. Weakly supervised discriminative models are able to improve
over the label models by learning complex interactions in the data,
making them more suited to handle complex decision
boundaries. Our results further experimentally confirm these findings,
as shown with the performance increase over label model predictions
discussed in section \ref{sec:results}, especially in the case of difficult
categories where heuristics are not sufficient.}

\begin{figure}[ht]
	\centering
	\includegraphics[scale=1]{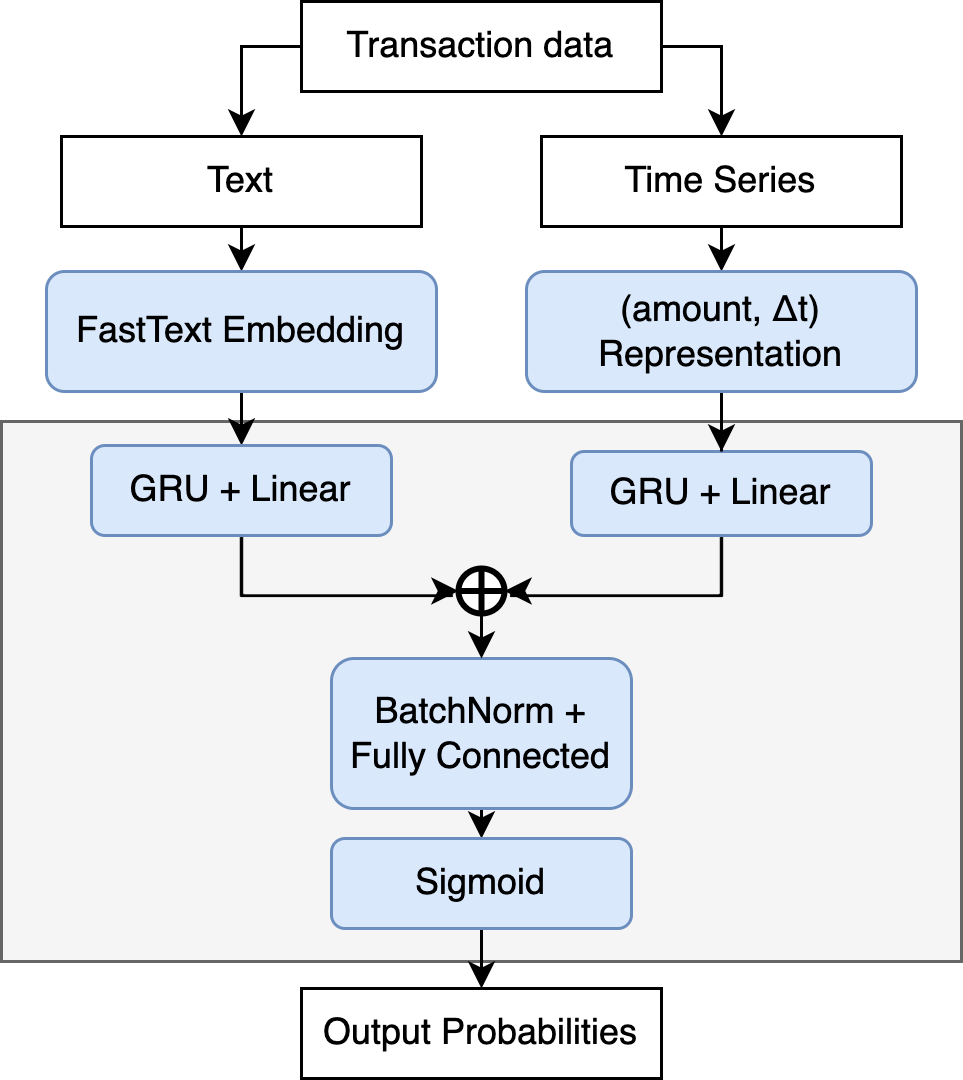}
	\caption{Discriminative Model Overview.}
	\label{fig:discr_overview}
\end{figure}

{As shown in Figure \ref{fig:discr_overview}, our DNN architecture consists of two
modalities, transaction text and transaction spending pattern. Both of
these inputs are sequences that we embed using a recurrent neural
network. These sequence embeddings are
concatenated before being passed into a fully connected multi layer
perceptron. Other transaction features or user features could be
concatenated in the input of the fully connected perceptron.
However, the methods described in this paper only include spending
patterns (time series) and text features.}

{Gated Recurrent Units (GRU) layers were chosen due to their relevance
to sequence classification, relative ease of implementation, and high
performance \cite{DBLP:journals/corr/ChungGCB14}. CNNs, RNNs, and LSTMs were experimented
with but did not perform better than GRUs and were less efficient. Other
techniques such as Transformers could be used, but could not in our
use case due to production limitations, namely limited training and
inference computing resources.}

{As a result of our preprocessing step, each group of transactions is
assigned a sparse two-dimensional time series of transaction amounts and
dates at which those transactions occur. By group definition, groups
have the same transaction text description, which is a sequence of
transaction words. Both of these sequences, spending pattern and
transaction description, are passed as input for our DNN.}

{First, we input the time series into a multi-layer bidirectional Gated
Recurrent Unit recurrent neural network  (GRU) to transform the sequence
into a fixed-dimension vector. To enhance computational speeds and
mitigate vanishing gradients, inputs pass through the GRU in a
variable-length packed format. After dimension reduction, the
GRU's final hidden states---combining forward and
backward directions---are used as the group's transaction sequence
embedding. By integrating transaction amounts and dates as features of
the same sequence, the time-series GRU embedding captures temporal
features of the transaction groups and the nature of transactions based
on their dollar amounts.}

{In parallel, the transaction text sequence is fed into another GRU to
embed the group's full transaction sentence. To match
the GRU's numerical format requirements, each word in
the sequence is first converted into word vectors using our transaction
FastText (\ref{sec:fasttext}) embedding. The text layers can then identify
meaningful correlations between the weak label and the transaction
sentence.}

{The text and spending pattern sequence embeddings are then concatenated
and passed into a multi-layer perceptron. Finally, the neural network
model outputs predicted probabilities of the transaction
group's category via a sigmoid layer. These predicted
probabilities serve as the final prediction of our transaction
classification system.}

{Multiple important factors were considered to enhance the results given
by training the discriminative model:}

\subparagraph{Training Data}

{A common train, validation, and testing split is used in our approach.
The same splits are used for both our label models and DNNs. }

{While label models are trained using the full training set, most
categories are highly imbalanced with some categories having an
occurrence of less than 1\% in the dataset. These imbalances hurt
the training of neural networks and it is necessary to use methods that can
handle large imbalances. Many approaches have been studied in the
literature from loss balancing, oversampling, undersampling, and many
more. We elected for the simple one-to-one random undersampling of our
training data using generated labels, giving the DNN a training set
consisting of equal parts positive weak labels and negative weak
labels.}

{Samples for which all labeling functions abstain are removed
prior to this undersampling from our DNN training dataset as performance
was severely affected with unlabeled samples. Since the label model is
able to generate probabilities for each category, we also experimented
with filtering out samples with low label model confidence. However,
this approach was not universal as some categories' performances were
lifted and some lowered. This could be due to two factors, firstly the
fact that our label models are not calibrated, partly from the
techniques used and the lack of annotated labels. Moreover, it is not
always beneficial to filter out low-confidence samples as being able to
learn how to classify them leads to the highest prediction power.
Consequently, the results and methodologies shown in this paper do not
incorporate this low-confidence filtering.}

\subparagraph{Weakly supervised training
dynamics}

{Various hyperparameters, such as learning rate, number of epochs,
training set size, and layer sizes, need to be refined for the models
to obtain acceptable performance. While it is possible to blindly run a
cross-validated hyperparameter search on a large search space to reduce
this need, this is not preferable, both due to the reduced number of
validation samples, required processing time, and the nature of neural
networks. Therefore, it's vital to carefully monitor
the training process and model outputs.}

{The training dynamics of our discriminative models revealed a
significant pattern. Initially, the discriminative model learns valuable
information to match input features to the weak labels, leading to an
improvement over the label model. However, as the training continues,
the discriminative model performance on the development set starts
declining, which is a clear indication of overfitting. Overfitting is
usually thought of as learning the training data too closely. In this
context, this problem is compounded as we also noticed overfitting to
the biases and errors of the label model. Instead of improving upon weak
labels, weakly supervised models are penalized in their training
objective when predicting differently from the weak labels. This can
negatively impact generalization ability and overall accuracy. }

{Naturally, we also found that capacity models like RNNs were more prone
to overfitting. Such models with a larger count of parameters could
achieve higher accuracies but also tended to overfit quickly, resulting
in significant accuracy variance between runs. To address these issues,
techniques such as early stopping, learning rate schedules, and
regularization played a critical role in preventing overfitting and
ensuring the discriminative model's robust performance.
In our use case, we used the Ray Tune library\cite{liaw2018tune}
to efficiently analyze loss and development set metrics over epochs and
training runs over hyperparameter search spaces. These insights helped
narrow out sets of hyperparameters that lead to robust model outputs.}

\subparagraph{Loss Functions}

{In the context of preventing overfitting to the weak labels, several
loss functions were experimented with throughout development. }

{While Binary Cross Entropy (BCE) and mean squared errors were initially
considered, BCE demonstrated better performance.
Poly-Loss \cite{leng2022polyloss} and Focal Loss \cite{lin2018focal}, two
extensions of BCE, were also tried but did not significantly enhance our
models' performance relative to the additional hyperparameters
required.}

{Given that weak labels are generated using a label model, the model's
targets are initially scores between 0 and 1 correlated to the
probability of a sample belonging to the positive class. As these labels
can be noisy and occasionally incorrect, we explored other noise-aware
losses to improve upon BCE for our use case. Out of the techniques
described in \cite{jiang2021named}, which revolved around weighing loss terms
by the confidence of the label model, we found that simply rounding the
weak labels targets to 0 or 1, a common approach in weak
supervision \cite{ratner2018training}, yielded the best results in our use case.}

{One possible explanation is that rounding the targets "forces" the
discriminative model to select a class with high confidence, whereas
keeping the probabilistic targets punishes the DNN for predicting a
different confidence to the label model. However, weighing losses by the
label model's confidence did not improve performance in our use case.
The absence of calibration of the label model due to the lack of ground
truth label and the sensitive nature of loss respective to balance
could explain this. Additionally, other methods such as Focal Loss \cite{lin2018focal}
demonstrate that under-weighing low-confidence samples is not always
beneficial, as learning how to separate difficult examples improves
decision power.}

\subsection{System Architecture}

{Apart from the efforts devoted to constructing the initial model
version, there are additional considerations that become
important once the system begins to operate in a production environment.
These considerations are particularly relevant due to the real-time
demands and the sensitive nature of the data processed. Training runs,
together with their artifacts and logs, need to be stored for
troubleshooting purposes as well as
regulatory compliance. Within a real-time setting, a continuous flow of
data from upstream systems must be incorporated into the engine for
inference while also being retained for future retraining. Parity
between training and inference is also crucial, and both need to run in
a distributed fashion as several workers need to coordinate, directly or
indirectly, to keep up with the demand. Finally, the
system's ability to seamlessly expand to accommodate new
classifications is a key consideration.}

\subsubsection{Transform design pattern}

{One of the key design patterns used to attain parity between training
and inference, as well as reproducibility, is the transform design
pattern \cite{Lakshmanan2020mlops}.}

{The core principle of the transform design pattern involves maintaining a
distinct separation between the raw data, features, and transformations.
This separation helps to maintain control over elements common to
various stages in the model's life cycle, such as
training, serving, and retraining. Additionally, it enables different
runs to be easily reproducible.}

{For instance, raw input data needs to be converted into appropriate
model features. This process is done in different ways depending on
the specific use case such as normalizing numerical features, encoding
text, or rescaling images. It is essential that the same
transformation is applied during training and inference. The transform
pattern achieves this by explicitly storing the transformation
that maps raw data to model features, and ensuring that it is
deterministic. In practice, this can be achieved by extracting any
parameters needed during training time. These parameters are
encapsulated within a transform object, and the object is stored so that
the precise same transformation can be applied later. In the case of
normalizing a numerical column, the mean and variance are computed
during the training stage and stored as an artifact to be used in
the inference preprocessing step. A simple but powerful example of the
transform pattern is the `fit \& transform` interface present in Machine
Learning libraries like scikit-learn.}

\subsubsection{Metadata Storage}

{Metadata refers broadly to information associated with each training
run used for the purposes of reproducibility, debugging,
monitoring, and compliance. Examples of metadata include the unique
identifier of the code used in a given training run, its start and end
time and log messages, as well as artifacts produced, such as actual
model weights or pre-trained embeddings.}

{Metadata is stored and organized using MLFlow. The first step at
every training run is to register it in MLFlow, mark it as started, and
assign it a unique ID, which can later be used to analyze the runs and
their results.}

{As the training progresses through multiple low-level components, the
internal state (which might be learned in some way by the component, for
example in the case of extracting a list of the most common words, or just
loaded from the start from some other source, such as a list of names)
of each is marshaled and stored as an independent artifact in MLFlow.
Once training is complete, MLFlow marks the runs as such, and if a model's
performance is increased, creates a new model version.}

{When inference starts, the best model version is fetched and all
artifacts associated with it are downloaded and loaded into memory. The
git hash of the current code version is also checked against the git
hash stored during training. This ensures that both code and artifacts
are identical to the training's.}

\subsubsection{Real-Time Inference}

{Raw transaction data is streamed into the pipeline from Apache Kafka.
Various events, such as the sign-up of a new company, new fields being filled by
users, or new transactions, are emitted from upstream services and
stored in different Kafka topics. Particularly, events corresponding to new
transactions are very frequent. Since our system needs
to be real-time, the naive approach would be to generate inferences for
individual transactions as soon as they arrive at the Kafka topic.
However, this is unfeasible, in part due to the high frequency of new data,
and in part due to the necessity to group transactions belonging to
each user in the model inference.}

{To circumvent this, KSQL, a streaming SQL engine for Apache Kafka, is
used to batch transactions and re-emit them as a single event consisting
of a single batch, each containing multiple transactions. The batches
are built either when a predetermined set amount of transactions have
been observed since the last batch in the original topic, or when a
predetermined time threshold has passed since the last batch. Lowering
these parameters puts more stress on computing resources. As such, they
are tuned accordingly to set the compromise between the compute resources
used by our microservice, and the rapidity of the system.}

{While the main entities used by the transaction categorization system
are transactions, inference also makes use of other data associated with
the company. Each type of data arrives in separate Kafka topics, and the
ordering of events is not guaranteed to be causally consistent. For a
variety of reasons, transaction events may arrive at the topic before
the event associated with the company sign-up, for example. Moreover,
business rules dictate the minimum amount of information and other
criteria required for each company for inferences to be generated.
Checking for these criteria using Kafka is however not practical,  as
they require grouping and aggregating statements over the events, which
cannot be done efficiently in the streaming context.}

{To efficiently address these challenges, we created a component called
a ``watcher''. Its job is to watch all important topics, store events
in the local disk, and decide whether there is enough data for any given
company. The watcher uses one file per company, which stores all events
associated with that company. Using local files allows much faster
read/write, lower infrastructure costs and maintenance overhead, as well
as being immune to network partitions. To avoid having to rebuild the
cache on startup, cache files are written to a mounted volume so they
are not affected by Kubernetes pods restarts.}

{Finally, once the watcher tags a company as ready for inference, the
data is fed into the inference pipeline which mirrors the same
sequences of transformations as the training pipeline.  After obtaining
the predicted probabilities of each categorization model, the final
output of the pipeline is packaged in the form of an event that is
emitted back to Kafka. The event contains the model type and version
used, along with the transaction id and the probability assigned to it
by the model. Downstream consumers can then collect these predictions
and do further work as needed by each use case. }

\subsubsection{Deployment}

{The training process takes place in Kubernetes. Each pipeline step
has a precise set of dependencies and outputs which are used to create a
directed acyclic graph (DAG). Kubeflow Pipelines is then used
to compile the execution DAG associated with the pipeline into an
Argo workflow. This allows the automatic parallel execution of
independent components as individual Kubernetes pods and for
independent scaling based on each component's resource
(CPUs, GPUs, and memory) requirements.}

{The intermediate and final outputs of the training pipeline, their logs,
 and other debugging information associated with each component, are
stored in a distributed object storage such as S3 or Minio. By doing so,
components can read their input independently of the particular
underlying server they happen to be assigned to.}

{To ensure reproducibility and the ability to trace problems, logs and
artifacts generated as part of Argo's execution process, 
such as the components' outputs,
are preserved and associated with a unique run in the
metadata storage, enabling tracking of data lineage, performance
metrics, and hyperparameters throughout the system. }

\subsubsection{New Task Expandability}

{The system architecture is designed for scalability, facilitating the
addition of new classification tasks and their corresponding models.
Incorporating a new category task requires data analysis, the
development of a small manually annotated sample set, the generation of
labeling functions, and the addition of the relevant parameters to the
system's settings file.}

\subparagraph{Labeling Functions}

{Formulating weak supervision rules is crucial when creating a new
model. These labeling functions emerge from an exploratory data analysis
of transactional data. Initial domain knowledge and data analysis can
provide basic rules for identifying a transaction description.
Heuristics can be derived from parsing the associated Plaid category and
identifying common spending patterns and text descriptions.}

{Notably, new tasks can leverage previous ones by incorporating their
labeling functions. For example, for a Restaurant expenses category,
labeling functions predicting recurring transactions can be negated. The
anchoring approach for text-based weak labeling described in \ref{sec:anchoring} further reduces iteration time.}

\subparagraph{Model Optimization}

{While the label-generating model and neural network methodologies
highlighted in this paper have proven effective for all transaction
classification tasks (nine to date), some task-specific parameters can
be adjusted. The class balance is pivotal in fitting the label model,
and proxies such as the Plaid category or educated estimations can help
assess it. Other model-specific parameters, including learning rate,
number of epochs, training set size, and layer sizes, should ideally be
fine-tuned for each task.}

\subparagraph{Iteration Duration}

{Overall, establishing a new categorization model necessitates
exploratory data analysis, development of labeling functions,
 annotation of a compact dataset for testing, and hyperparameter tuning.
After the initial cost of implementing the proposed system, the basic
setup for a new classification model can be completed in as little as a
week.}

\section{Results}
\label{sec:results}

{In order to evaluate the performance of our methods, we showcase
classification metrics of our label models and discriminative models for
nine classification tasks. These results are compared to a baseline
consisting of predicted categorizations given by the Plaid API, the
current market-leading provider of transaction information.}

{As previously mentioned, our manually annotated testing sets have
shortcomings such as their relatively small size or potential bias
towards transactions with clearer descriptions and patterns. To give an accurate
estimate of performance, we ensured all our
testing sets have at least one hundred positive samples and one hundred
negative samples, a difficult task for categories that have as low as
0.5\% positive rate. Moreover, we used a train-development-testing split
to assess performance on unseen data.}

{Model performance is showcased using the balanced accuracy metric.
Balanced accuracy is an intuitive way of assessing the performance for
imbalanced problems, especially when the performance of both classes is
important. It is computed by averaging the accuracy of the classifier on
a set composed of positive samples and another set composed of negative
samples. Balanced accuracy ensures models are not simply predicting one
class over the over, with a score of 0.5 corresponding to completely
random (worst) predictions, and 1 for a perfect classifier. 
Results for recall and a more detailed discussion of our model's performance vs. Plaid can be found in appendix \ref{app:recall}.}

{As showcased in appendix \ref{app:ablation}, the performance of models can
be relatively varied for different seeds. This variance can positively or negatively affect
our given performance metrics. In order to give a realistic estimate of
the discriminative model performance, we show the result of the 10th
best performing model on the validation set out of fifty training runs for each task.}

% Increase row height by 10%
\renewcommand{\arraystretch}{1.2}
% Increase column separation by 10%
\setlength{\tabcolsep}{1.2\tabcolsep}

\begin{table}[ht]
\centering
\scalebox{1.00}{
\begin{tabular}{|c|c|c|c|c|}
\hline
%\textbf{Models} & \textbf{Plaid API (Baseline)} & \textbf{Label Model} & \textbf{DNN Model} & \textbf{Improvement over Plaid} \\
\multirow{2}{*}{\textbf{Models}} & \multirow{2}{*}{\textbf{Plaid API}} & \textbf{Label} & \textbf{DNN} & \textbf{Improvement} \\
 & & \textbf{Model} & \textbf{Model} &  \textbf{Over Plaid} \\
\hline
Ads and Marketing & 0.66 & 0.72 & \underline{\textbf{0.93}} & +0.27 \\
\hline
Insufficient Funds Fees & 0.92 & 0.97 & \underline{\textbf{0.99}} & +0.07 \\
\hline
Wages and Payroll & 0.90 & \underline{\textbf{0.92}} & 0.60 & +0.02 \\
\hline
Personal & \textit{None} & \underline{\textbf{0.91}} & 0.87 & \textit{N/A} \\
\hline
Recurring & \textit{None} & 0.77 & \underline{\textbf{0.86}} & \textit{N/A} \\
\hline
Rent & 0.53 & 0.57 & \underline{\textbf{0.76}} & +0.23 \\
\hline
Revenue & \textit{None} & \underline{\textbf{0.90}} & 0.74 & \textit{N/A} \\
\hline
Telecoms & \underline{\textbf{0.98}} & 0.97 & 0.92 & -0.01 \\
\hline
Utilities & 0.71 & \underline{\textbf{0.93}} & 0.89 & +0.22 \\
\hline
\end{tabular}}
\caption{Balanced accuracy for benchmark Plaid categories,
Label Model and DNN Model.}
\label{table:results}
\end{table}

%CHECK THAT THIS DOES NOT BREAK STUFF
\renewcommand{\arraystretch}{1}
\setlength{\tabcolsep}{\tabcolsep}

{The results of the Plaid categorizations can vary widely. The
Plaid Telecoms category performs extremely well, as is easily understood
since transaction descriptions are constrained to a finite list of
telecommunication company names and a few other keywords. Insufficient
Funds Fees is its next best category, but Rent and Marketing have very
low performance. In the case of Marketing, it may be that there is a
lack of time spent on refining the categorization, but in the case of
Rent, classification is inherently challenging. Finally, the Plaid
categories used as baselines are limited and cannot be extended to some
tasks we created such as Revenue, Personal (non-business) expenses, or
Recurring expenses. }

{The Label Model was carefully tuned for each task and was able to
outperform or match Plaid for every benchmark. For tasks categorizing a
specific list of merchants and transactions, precise labeling functions
could be tailored to achieve very high performance, such as in Wages
and Payroll with 92\% or Utilities with 93\%. As shown with these
examples, the Label Model performed particularly well for tasks that are
easily tractable, even surpassing the weakly supervised neural networks.}

{However, for more difficult tasks, such as Rent, Recurring, and
Marketing, the discriminative weakly supervised DNN shows a
clear performance increase over the Label Model. In particular, the Advertising model was able to achieve 93\% balanced accuracy, 27 points over its Label Model, and the Rent
DNN was able to achieve 76\% balanced accuracy, 19 points over its Label
Model, which was much higher than we anticipated. The broad Recurring
task also shows a 9 points improvement over its label model.}

{The results in Table \ref{table:results} demonstrate our proposed methods can
achieve accurate transaction classification, with an average balanced
accuracy of 91\% across all tasks. While matching or improving the
performance on the three categories where Plaid has over 90\% balanced
accuracy, our proposed method outperforms by more than 20 points the
other three Plaid categorizations that have lower balanced accuracy. In
the three instances of label models with less than 90\% balanced
accuracy, training the discriminative model yields on average +16
accuracy over weak labels. We also show strong results on three new
categories that are not part of the Plaid schema, proving the
flexibility of our proposed method beyond Plaid and on a total of nine
classification tasks. We believe these results further demonstrate the
power of weak supervision in practical use cases and prove our
proposed solution can achieve accurate and scalable bank transaction
classifications, which can, in turn, unlock exciting applications in the
financial world. }

\section{Future Work}

{A series of improvements and advancements could greatly enhance the
efficacy and efficiency of our model. Beyond improving the transaction
classification techniques used, many exciting applications can be
developed using accurate transaction classifiers. This chapter outlines
these prospective areas for such refinements and developments.}

\subsection{Technical improvements}

\subparagraph{Active learning}

{Active learning could be explored in future work to further improve the
performance of our model and enhance its adaptability to new data.
Active learning is an approach that iteratively detects and flags for
annotation the most valuable unlabeled samples in datasets, allowing
models to learn effectively from fewer annotations. This approach can 
drastically reduce the annotation burden and facilitates a more efficient
learning process. We would also like to explore a combination of active
learning for both strong labels and also weak labels and identify ways
to detect where new labeling functions should be created.}

\subparagraph{Transfer learning}

{Transaction language pretraining was accomplished using FastText to
learn meaningful representations of transaction descriptions. This work
could be refined but also extended to the time series component of our
model. Moreover, most classification tasks have some underlying
correlation with the recurring model. For instance, rent payments are
necessarily recurrent and restaurant payments are not. The recurring
model also has the nice property of being more balanced, and thus
trained with considerably more samples than other tasks due to our
undersampling approach. Considering these dependencies and properties,
it is likely that fine-tuning the recurring model for the other more
specific classification tasks could lead to performance improvements.}

\subparagraph{Label model improvements}

{The probabilistic model used to generate labels in this paper relied on
the open-source Snorkel library. Sadly, the Snorkel open-source work has
been unofficially on hold since 2020. The current implementation of the
Snorkel label model does not take into consideration conditional
dependencies between labeling functions, utilizing a naive Bayes
assumption, and does not take abstentions of labeling functions in its
posterior calculation. We have experimentally noticed that while Snorkel
can separate positive and negative samples in a statistically powerful
manner, the assigned probabilities are not always calibrated, which can
lead to bad results. In other experiments, as noted in
\cite{ratner2018training}, using information gained from abstentions in the
posterior calculation also improved label model results.}

{Beyond Snorkel, we believe many improvements can be brought to our
weakly supervised approach, including more efficient ways of
constructing labeling functions, using data-aware label models,
calibration of label models, improved generative models for label
models, and weakly supervised aware discriminative model training
techniques. }

\subparagraph{Discriminative model improvements}

{While our proposed dual GRU architecture showed good results, this
approach is only one out of many that could be successful in this use
case. Our transaction FastText embedding is only tangled with the weak
labels during the DNN training stage, which is severely undersampled in
the case of low balance classifications. Techniques that could learn
from the entire weakly supervised data, such as Transformers or
conditional generative models, could be better suited and more robust in
the context of highly imbalanced datasets like our use case.}

\subparagraph{Training techniques}

{Current discriminative training techniques are entirely classical and
adapting training to weak supervision could provide significant
benefits. Traditional overfitting issues are exacerbated in weak
supervision as models tend to learn the weak targets over time,
which stops the progression of discriminative models during training.
Investigating ways to prevent or delay this weak-label overfitting could
yield significant improvements. Additionally, while our models were
initially trained using Stochastic Gradient Descent, experiments with the
AdamW \cite{loshchilov2019decoupled} optimizer show an increase of +6\% performance for the advertising model and +2\% for the recurring model suggesting that a
deeper understanding of training dynamics in the context of weak-label
noise could constitute exciting research.}

\subsection{Use Cases and Applications}

{The financial world is built on transactions and being able to classify
them can have a wide array of applications. In our research, we have
focused on bank transactions, specifically those related to businesses.
The categorization of these transactions can provide valuable insights
and support the development of sophisticated financial services, such as
credit risk assessment, financial health reporting, fraud detection, and
many others. }

\subparagraph{Credit Risk}

{The traditional credit scoring model relies heavily on positive credit
history, making it challenging for thin-file users or those with past
financial mishaps to access and build credit history. These credit
scoring systems also give a delayed view of financial health due to
their reliance on credit products and third parties. Transactions can
help give powerful and real-time insights into user spending behavior.
Regular payments such as telecom, utilities, rent, mortgage, or loan
payments can be used as low-latency indicators of their ability to repay
debt. For categories such as utilities and telecom, spending amounts are
lower and add lower weights to level 4 credit lines. On the other hand,
categories such as rent and loans are stronger indicators of financial
stability and carry more weight as level 3 credit lines. Some of the
most relevant transaction categories that could be added for this
purpose on top of the nine implemented in this paper could be mortgages,
insurance payments, car loans, and subscription services.}

{In the context of business accounts, classifying personal transactions,
as opposed to business spending, also has inherent value. The mixing of
business and personal expenses on a single account is known as
commingling and is taken into account by banks and for credit risk.
Utilizing business accounts for personal expenses is against terms set
by banks and can lead to rewards reversal or account closure. It is also
another indicator of financial health that can be used to identify
responsible business spending behaviors. Other categories such as
advertising spend can also be used to determine the maturity and
financial health of a business.}

\subparagraph{User-Facing Financial Insights}

{Presenting financial insights to users can foster engagement and
teach how to improve financial health. Just as the credit risk
factors described above can help credit issuers assess risk, they can
also be reported directly to the user to help them better understand and
improve their financial situation. Utilizing transactions to assess risk
in this manner could fix the current black-box nature of credit scoring.
Metrics of health and risk can be highly valuable in matching users
with appropriate credit products. Transaction classification is also
the backbone of cash-back and rewards. More recently, the surge of
subscriptions and forgotten free tree trials has led to the development
of many products aiming to help users manage these expenses}

{Using monitoring and educational insights, users can be helped to
attain better credit scores and be matched with better credit offers. One
idea is to use gamification and award badges and rewards to users who display
healthy financial behavior. This approach could help
reduce credit risk and establish a basis for client-company interaction,
helping build a financially responsible and engaged user base. Using an
active learning and crowdsourcing annotation approach, such users could
also help the product and its annotations, especially for misclassified
transactions. Curated sets of transaction labels are hard to come by,
and the quality of the platform could be improved and expanded by
obtaining such annotations from users.}

{For businesses, benchmarks and financial spending metrics comparisons to
fellow users of the same industry can help businesses direct their
finances and spending.  With the ability to categorize recurring
transactions, users could also sign up for incident monitoring
to predict and prepare upcoming recurring expenses. This includes
pushing reminders that help reduce the occurrence of insufficient funds
or lateness incidents. Moreover, thin-file users could also sign up to
establish a first credit line by promoting their positive financial
transaction history.}

\subparagraph{User-level embedding}

{User-level financial information embeddings, such as the Client2Vec
approach described in \cite{baldassini2018client2vec}, are compressed
representations of users that can be used in many tasks such as
transaction-based user segmentation, fraud detection, and product
targeting. Such approaches help perform user-level tasks more accurately and efficiently.
The learned representation would further
surface deep associations between user behaviors, possibly lifting out
user archetypes and financial health. Representative segment identification is additionally
key for targeted marketing and strategizing.}

\section{Conclusion}
% $^{\ref{sec:results}}$
% $^{\ref{sec:fasttext}}$
% $^{\ref{sec:weak}}$
% $^{\ref{sec:dnn}}$
{In conclusion, we have presented a scalable approach for bank transaction classification. We showcase how to bypass data annotation by building upon domain knowledge which offers the usage of state-of-the-art classification tools for large and varied unlabeled datasets. Our solution is both cost-effective and superior to existing methods like Plaid in terms of accuracy. Moreover, it is capable of supporting novel and composite tasks within a fast implementation timeframe. This demonstrates the effectiveness of combining embeddings, probabilistic labels built from anchoring and heuristics, and multimodal deep neural networks for accurately categorizing transactions. We believe our work opens promising avenues for real-world uses such as credit risk assessment and financial health monitoring and provides a practical example of weak supervision techniques for product applications.}

\section{Acknowledgements}

{We would like to thank Matias Benitez for his many early contributions
to the project, including an initial proof of concept written in
TensorFlow and Keras.}

% BIBLIOGRAPH
\printbibliography

@article{schmarje2021survey,
  title={A survey on semi-, self-and unsupervised learning for image classification},
  author={Schmarje, Lars and Santarossa, Monty and Schr{\"o}der, Simon-Martin and Koch, Reinhard},
  journal={IEEE Access},
  volume={9},
  pages={82146--82168},
  year={2021},
  publisher={IEEE}
}

@misc{polars,
  author = {{Polars Contributors}},
  title = {Polars: DataFrame library implemented in Rust},
  year = {2023},
  publisher = {GitHub},
  journal = {GitHub repository},
  howpublished = {\url{https://github.com/pola-rs/polars}},
  note = {Required Rust version >=1.62}
}

@misc{mcinnes2020umap,
      title={UMAP: Uniform Manifold Approximation and Projection for Dimension Reduction}, 
      author={Leland McInnes and John Healy and James Melville},
      year={2020},
      eprint={1802.03426},
      archivePrefix={arXiv},
      primaryClass={stat.ML}
}

@misc{torchtext,
  title = {TorchText Library},
  author = {Torch},
  year = {2023},
  note = {Available at: \url{https://pytorch.org/text/}}
}

@article{crichton2017ner,
    title={A neural network multi-task learning approach to biomedical named entity recognition},
    author={Gamal K. O. Crichton and Sampo Pyysalo and Billy Chiu and Anna Korhonen},
    journal={BMC Bioinformatics},
    year={2017},
    volume={18},
    number={1},
    doi={10.1186/s12859-017-1776-8}
}

@book{Lakshmanan2020mlops,
    title = {Machine Learning Design Patterns: Solutions to Common Challenges in Data Preparation, Model Building, and MLOps},
    author = {Valliappa Lakshmanan and Sara Robinson and Michael Munn},
    year = {2020},
    publisher = {O'Reilly Media, Inc.}
}

@misc{liaw2018tune,
      title={Tune: A Research Platform for Distributed Model Selection and Training}, 
      author={Richard Liaw and Eric Liang and Robert Nishihara and Philipp Moritz and Joseph E. Gonzalez and Ion Stoica},
      year={2018},
      eprint={1807.05118},
      archivePrefix={arXiv},
      primaryClass={cs.LG}
}

@misc{ratner2017data,
      title={Data Programming: Creating Large Training Sets, Quickly}, 
      author={Alexander Ratner and Christopher De Sa and Sen Wu and Daniel Selsam and Christopher Ré},
      year={2017},
      eprint={1605.07723},
      archivePrefix={arXiv},
      primaryClass={stat.ML}
}

@misc{Plaid2023Transactions,
  author       = {PlaidInc.},
  title        = {Transactions},
  year         = {2023},
  url          = {https://plaid.com/docs/transactions/}
}

@inproceedings{olivier2018, 
 author = {Oliver, Avital and Odena, Augustus and Raffel, Colin A and Cubuk, Ekin Dogus and Goodfellow, Ian},
 booktitle = {Advances in Neural Information Processing Systems},
 editor = {S. Bengio and H. Wallach and H. Larochelle and K. Grauman and N. Cesa-Bianchi and R. Garnett},
 pages = {},
 publisher = {Curran Associates, Inc.},
 title = {Realistic Evaluation of Deep Semi-Supervised Learning Algorithms},
 url = {https://proceedings.neurips.cc/paper_files/paper/2018/file/c1fea270c48e8079d8ddf7d06d26ab52-Paper.pdf},
 volume = {31},
 year = {2018}
}

@misc{loshchilov2019decoupled,
      title={Decoupled Weight Decay Regularization}, 
      author={Ilya Loshchilov and Frank Hutter},
      year={2019},
      eprint={1711.05101},
      archivePrefix={arXiv},
      primaryClass={cs.LG}
}

@misc{xu2020iterative,
      title={Iterative Pseudo-Labeling for Speech Recognition}, 
      author={Qiantong Xu and Tatiana Likhomanenko and Jacob Kahn and Awni Hannun and Gabriel Synnaeve and Ronan Collobert},
      year={2020},
      eprint={2005.09267},
      archivePrefix={arXiv},
      primaryClass={cs.CL}
}

@misc{mikolov2013efficient,
      title={Efficient Estimation of Word Representations in Vector Space}, 
      author={Tomas Mikolov and Kai Chen and Greg Corrado and Jeffrey Dean},
      year={2013},
      eprint={1301.3781},
      archivePrefix={arXiv},
      primaryClass={cs.CL}
}

@misc{leng2022polyloss,
      title={PolyLoss: A Polynomial Expansion Perspective of Classification Loss Functions}, 
      author={Zhaoqi Leng and Mingxing Tan and Chenxi Liu and Ekin Dogus Cubuk and Xiaojie Shi and Shuyang Cheng and Dragomir Anguelov},
      year={2022},
      eprint={2204.12511},
      archivePrefix={arXiv},
      primaryClass={cs.CV}
}

@misc{lin2018focal,
      title={Focal Loss for Dense Object Detection}, 
      author={Tsung-Yi Lin and Priya Goyal and Ross Girshick and Kaiming He and Piotr Dollár},
      year={2018},
      eprint={1708.02002},
      archivePrefix={arXiv},
      primaryClass={cs.CV}
}

@misc{jiang2021named,
      title={Named Entity Recognition with Small Strongly Labeled and Large Weakly Labeled Data}, 
      author={Haoming Jiang and Danqing Zhang and Tianyu Cao and Bing Yin and Tuo Zhao},
      year={2021},
      eprint={2106.08977},
      archivePrefix={arXiv},
      primaryClass={cs.CL}
}

@article{lv2019,
author = {Fang Lv and Junheng Huang and Wei Wang and Yuliang Wei and Yunxiao Sun and Bailing Wang},
title = {A two-route CNN model for bank account classification with heterogeneous data},
journal = {PLoS ONE},
volume = {14},
number = {8},
pages = {e0220631},
year = {2019},
url = {https://journals.plos.org/plosone/article?id=10.1371/journal.pone.0220631},
doi = {10.1371/journal.pone.0220631},
eprinttype = {PloS ONE},
eprint = {10.1371/journal.pone.0220631}
}

@article{kotios2022,
author = {Dimitrios Kotios and
Georgios Makridis and
Georgios Fatouros and
Dimosthenis Kyriazis},
title = {Deep learning enhancing banking services: a hybrid transaction classification and cash flow prediction approach},
journal = {Journal of Big Data},
volume = {9},
pages = {100},
year = {2022},
url = {https://journalofbigdata.springeropen.com/articles/10.1186/s40537-022-00651-x},
doi = {10.1186/s40537-022-00651-x},
eprinttype = {Journal of Big Data},
eprint = {10.1186/s40537-022-00651-x}
}

@inproceedings{zhang2021,
author = {Chen Zhang and
Qian Wang and
Ting Liu and
Xuan Lu and
Jialin Hong and
Bin Han and
Chen Gong},
title = {Fraud Detection under Multi-Sourced Extremely Noisy Annotations},
booktitle = {Proceedings of the 30th ACM International Conference on Information {\&} Knowledge Management},
year = {2021},
organization = {CIKM '21}
}

@article{radford2022,
author = {Alec Radford and
Jong Wook Kim and
Tianshi Xu and
Greg Brockman and
Cullen McLeavey and
Ilya Sutskever},
title = {Robust Speech Recognition via Large-Scale Weak Supervision},
journal = {arXiv preprint arXiv:2212.04356},
year = {2022},
url = {https://arxiv.org/abs/2212.04356},
eprinttype = {arXiv},
eprint = {2212.04356}
}

@article{chen2022,
author = {Meng Fang Chen and others},
title = {Shoring Up the Foundations: Fusing Model Embeddings and Weak Supervision},
journal = {arXiv preprint arXiv:2203.13270},
year = {2022},
url = {https://arxiv.org/pdf/2203.13270.pdf},
eprinttype = {arXiv},
eprint = {2203.13270}
}

@misc{statista2022,
author = {Statista},
title = {Number of QSR, FSR, chain and independent restaurants in the U.S.},
year = {2022},
url = {https://www.statista.com/statistics/244616/number-of-qsr-fsr-chain-independent-restaurants-in-the-us/},
note = {[Graph]. In Statista. Retrieved April 30, 2023}
}

@book{ousterhout2018,
author = {John Ousterhout},
title = {A Philosophy of Software Design},
publisher = {Yale University Press},
address = {New Haven, CT},
year = {2018}
}

@misc{baldassini2018client2vec,
      title={client2vec: Towards Systematic Baselines for Banking Applications}, 
      author={Leonardo Baldassini and Jose Antonio Rodríguez Serrano},
      year={2018},
      eprint={1802.04198},
      archivePrefix={arXiv},
      primaryClass={stat.ML}
}

@INPROCEEDINGS{athiwaratkun2017,
  author={Athiwaratkun, Ben and Stokes, Jack W.},
  booktitle={2017 IEEE International Conference on Acoustics, Speech and Signal Processing (ICASSP)}, 
  title={Malware classification with LSTM and GRU language models and a character-level CNN}, 
  year={2017},
  volume={},
  number={},
  pages={2482-2486},
  doi={10.1109/ICASSP.2017.7952603}
}

@article{salur2020,
author = {Salur, Mehmet and Aydin, Ilhan},
year = {2020},
month = {04},
pages = {58080 - 58093},
title = {A Novel Hybrid Deep Learning Model for Sentiment Classification},
volume = {8},
journal = {IEEE Access},
doi = {10.1109/ACCESS.2020.2982538}
}

@article{ratnersnorkel,
  author       = {Alexander Ratner and
                  Stephen H. Bach and
                  Henry R. Ehrenberg and
                  Jason Alan Fries and
                  Sen Wu and
                  Christopher R{\'{e}}},
  title        = {Snorkel: Rapid Training Data Creation with Weak Supervision},
  journal      = {CoRR},
  volume       = {abs/1711.10160},
  year         = {2017},
  url          = {http://arxiv.org/abs/1711.10160},
  eprinttype    = {arXiv},
  eprint       = {1711.10160},
  bibsource    = {dblp computer science bibliography, https://dblp.org}
}

@misc{ratner2018training,
  	title={Training Complex Models with Multi-Task Weak Supervision},
  	author={Alexander Ratner and Braden Hancock and Jared Dunnmon and Frederic Sala and Shreyash Pandey and Christopher Ré},
  	year={2018},
  	eprint={1810.02840},
  	archivePrefix={arXiv},
  	primaryClass={stat.ML}
}

@article{DBLP:journals/corr/ChungGCB14,
  author       = {Junyoung Chung and
                  {\c{C}}aglar G{\"{u}}l{\c{c}}ehre and
                  KyungHyun Cho and
                  Yoshua Bengio},
  title        = {Empirical Evaluation of Gated Recurrent Neural Networks on Sequence
                  Modeling},
  journal      = {CoRR},
  volume       = {abs/1412.3555},
  year         = {2014},
  url          = {http://arxiv.org/abs/1412.3555},
  eprinttype    = {arXiv},
  eprint       = {1412.3555},
  bibsource    = {dblp computer science bibliography, https://dblp.org}
}

@article{rehurek2011gensim,
  title={Gensim--python framework for vector space modelling},
  author={Rehurek, Radim and Sojka, Petr},
  journal={NLP Centre, Faculty of Informatics, Masaryk University, Brno, Czech Republic},
  volume={3},
  number={2},
  year={2011}
}

@article{joulin2016bag,
  title={Bag of Tricks for Efficient Text Classification},
  author={Joulin, Armand and Grave, Edouard and Bojanowski, Piotr and Mikolov, Tomas},
  journal={arXiv preprint arXiv:1607.01759},
  year={2016}
}

% APPENDIX
\newpage
\begin{appendices}
\appendix
\section{Training and Inference Flow}
Our transaction categorisation system has two major branches, one used for training and one used for inference. The two branches share many commonalities, including the data loading and preprocessing steps, shown below.

  \begin{figure}[ht!]
	\centering
	\includegraphics[scale=0.40]{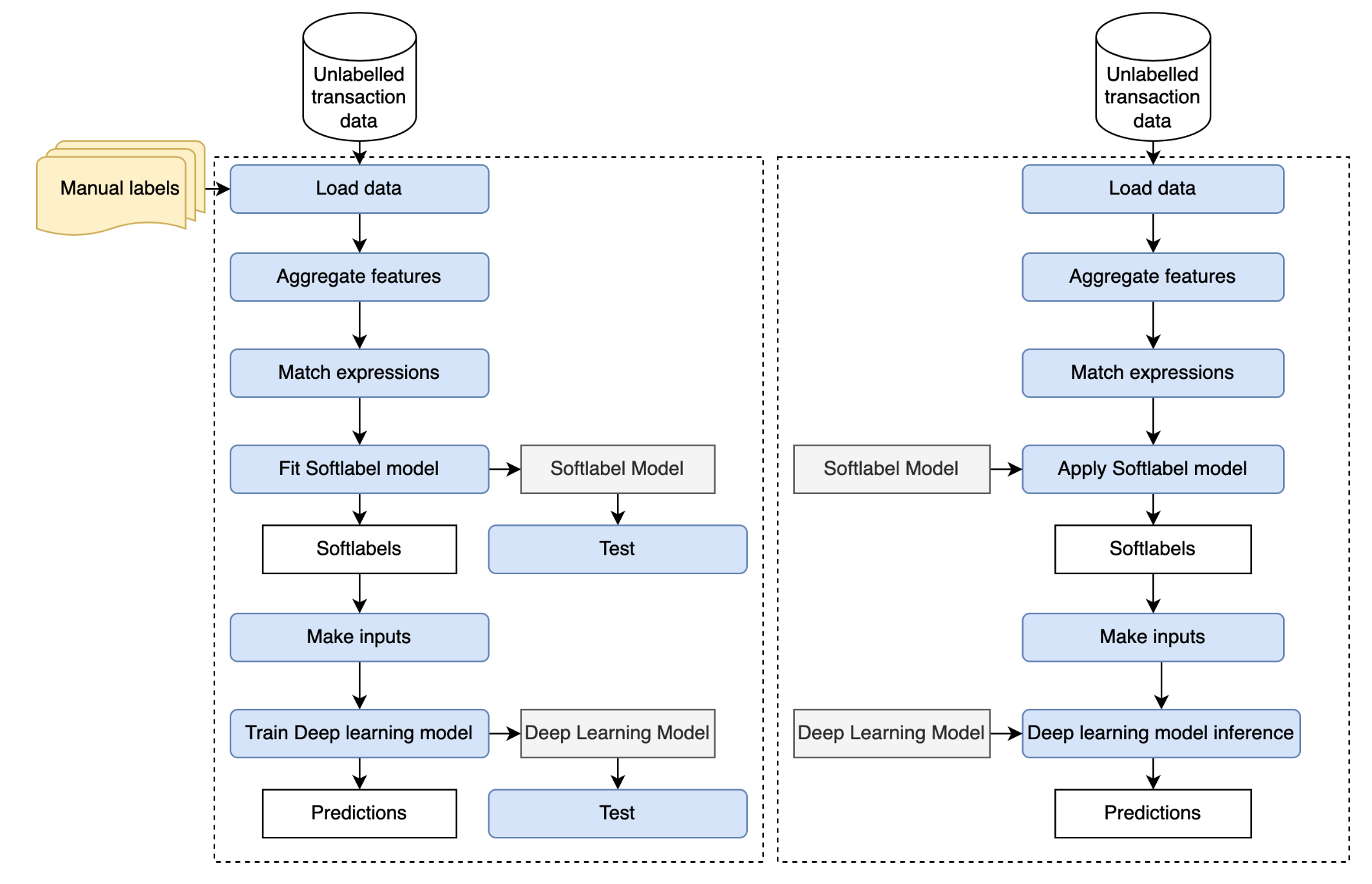}
	\caption{Categorisation model training and inference pipeline overview.}
	\label{fig:pipelines}
\end{figure}

The training pipeline loads data according to preset requirements, cleans and aggregates the data, with grouping as discussed in the paper. We perform text pattern matching in the step called “match expressions”. We use the polars\cite{polars} library to apply the regular expression match across a large dataset in a parallel and efficient fashion. The cleaned data and aggregate features are sent to the weak label model which uses labeling functions to fit a weak label model to create probabilistic labels. These are tested against the small amount of ground truth labels to determine the quality of the weak label model. The labels and preprocessed data are then converted to inputs for the deep learning model and the probabilistic labels are used as labels for supervised training. The resulting model is again tested against human annotated ground truth labels. The training pipeline produces two saved models, and their respective metrics tested against ground truth labels.

The inference pipeline mirrors the training pipeline in that it loads the models trained and saved in the training pipeline and performs inference on the loaded data. The data is preprocessed in the same way as in the training pipeline. This same pipeline is used to train any type of categorisation model for transaction categorisation.

\section{Discriminative Model Experiments and Ablation}
\label{app:ablation}
To optimize the classification of bank transaction data, we undertook a series of experiments and ablation studies which centered on the application of various discriminative models. In this section, we describe our exploratory journey and subsequent findings.

\paragraph{Off the shelf English corpus embeddings}
We explored various pre-trained English word vector embeddings from the torchtext\cite{torchtext} library, namely wiki-en-FastText (pretrained on Wikipedia), CharNGram and GloVe. Initial versions of our DNNs used these pretrained embeddings and both retraining or freezing embeddings was experimented with.

We found that the character based CharNGram embedding performed best compared to the pretrained FastText or GloVe word embeddings. FastText, also based on n-grams, also performed fairly well, reinforcing the intuition that character level representations are important for unstructured text such as transaction text.

\paragraph{Weakly supervised CNN character embedding}
While integrating English corpus embeddings enhanced performance as opposed to using sparse non-pre-trained weakly supervised embeddings, transaction text was observed to contain a significant number of novel tokens, including abbreviations, truncated words, custom tokens, and undecipherable tokens.

In order to further optimize the model, we utilized the hybrid model method as proposed by Salur et al\cite{salur2020}. This method combines word-level embeddings, such as Word2Vec and FastText, with different character-level deep learning embeddings (LSTM, GRU, CNN). Our best-performing model utilized CNNs for the character sequence and a bi-directional GRU embedding for the word sequence, concatenated with off-the-shelf FastText word vectors.

One key advantage of this approach is that the CNN model is able to learn unique character combinations and abbreviations of common words which the text embeddings may miss. However, the computational complexity of training these CNNs on large amounts of data rendered this approach impractical in our use-case. We ultimately favored retraining a FastText embedding on transactional text, which not only proved more efficient, but also resulted in the best performance.

\paragraph{Ablation Studies and Decision on Neural Network Architecture}
Our final choice of neural network architecture was informed by a series of ablation studies performed on the DNN architecture. We tested all possible combinations of three potential modalities: time series, CharNGram text embeddings, and 1D CNN character embeddings. Our findings were then compared to the final proposed neural network architecture which used transaction-corpus-trained FastText embeddings and transaction spending sequence. By running 50 random trials for each ablation, we compare the balanced accuracies mean, variance, and maximum for each architecture. The results of these studies are summarized in Table \ref{table:ablation} below.\footnote{NB: This study was constructed during an earlier phase of our project when SGD and different sets of hyperparameter were used compared to AdamW and our final hyperparameters used to construct the results in \ref{table:results}.}

\begin{table}[ht!!]
\begin{center}
\begin{tabular}{lcccccc}
\hline
 & \multicolumn{3}{c}{Utilities} & \multicolumn{3}{c}{Rent} \\
\cline{2-4} \cline{5-7}
 & mean & std & max & mean & std & max \\
\hline
Time Series & 0.531 & 0.040 & 0.601 & 0.634 & 0.062 & 0.747 \\
Text & 0.586 & 0.086 & 0.680 & 0.495 & 0.011 & 0.517 \\
Char & 0.542 & 0.055 & 0.640 & 0.512 & 0.022 & 0.552 \\
Time Series-Text & 0.550 & 0.045 & 0.680 & 0.628 & 0.066 & 0.730 \\
Time Series-Char & 0.531 & 0.048 & 0.685 & 0.652 & 0.073 & 0.793 \\
Text-Char & 0.608 & 0.087 & 0.742 & 0.504 & 0.016 & 0.540 \\
Time Series-Text-Char & 0.567 & 0.059 & 0.725 & 0.642 & 0.069 & \textbf{0.816} \\
Time Series-Text with pretrained FastText & \textbf{0.723} & 0.073 & \textbf{0.843} & \textbf{0.665} & 0.070 & 0.805 \\
\hline
\end{tabular}
\caption{Results for different ablation experiments.}
\label{table:ablation}
\end{center}
\end{table}

  \begin{figure}[ht!]
	\centering
	\includegraphics[scale=0.35]{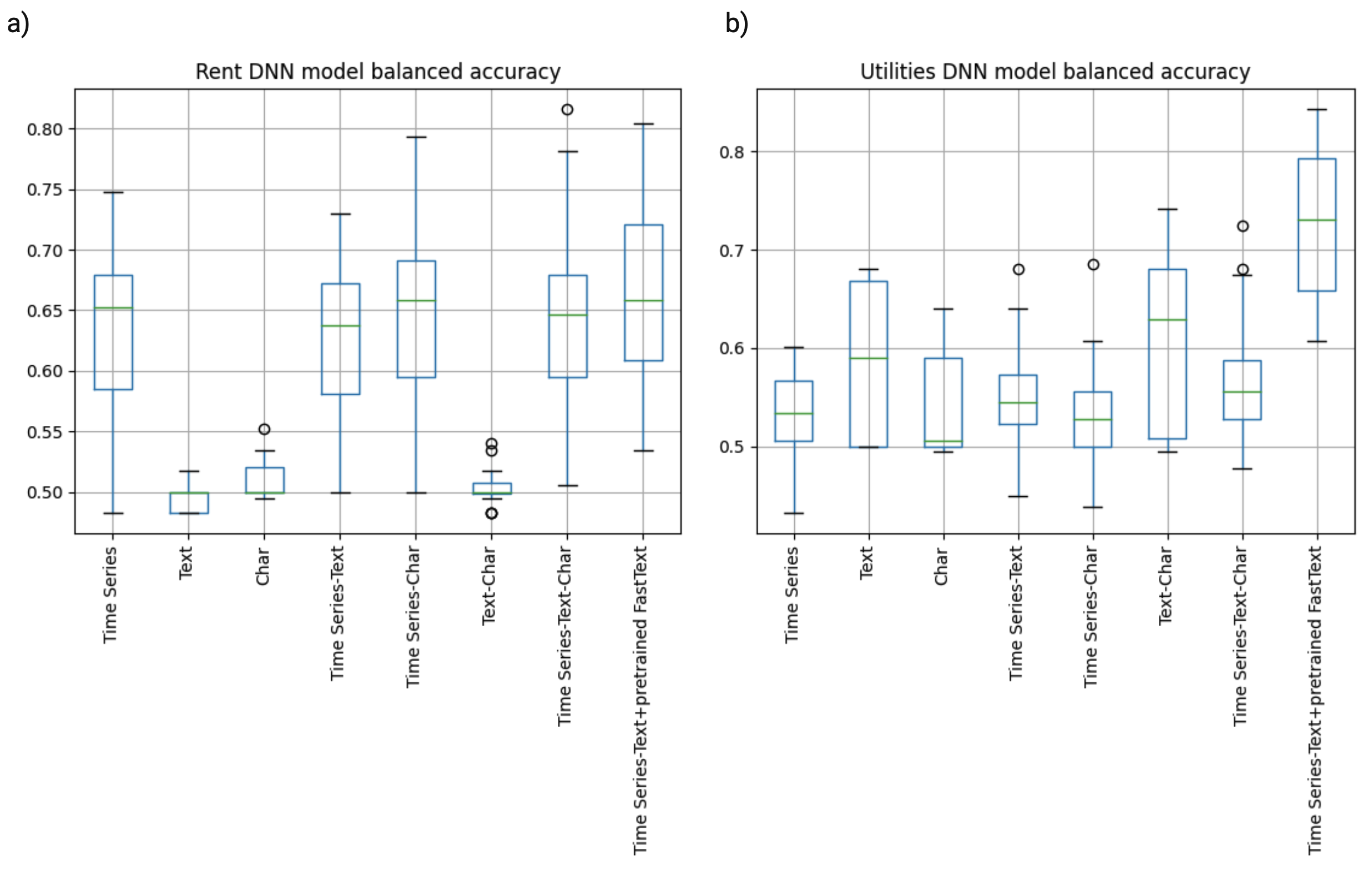}
	\caption{Ablation over 50 trials for a) rent and b) Utilities DNN models.}
	\label{fig:ablation}
\end{figure}

Notably, the performance of the different architectures varied widely depending on the classification task at hand. It was clear that some models relied more heavily on text while others were dependent on transaction amount history. However, after conducting experiments with training a FastText embedding on the transaction corpus, the TimeSeries+FastText approach consistently yielded better or similar results compared to previous best infrastructures for each of our nine classification tasks.

\paragraph{Conclusive Remarks}
The exploratory journey depicted in this appendix, involving various experiments and ablation studies, has led us to use the TimeSeries+FastText architecture for all tasks. This model, through its union of purpose-trained transaction text embeddings with transaction time series, has shown its capability to classify transaction patterns with efficiency and generality.

Ablation studies revealed the varying performance of each modality depending on the task at hand, highlighting the need for a multi-modal model infrastructure. Despite such variability, the TimeSeries+FastText approach consistently offered superior or comparable results across all tasks, reinforcing its selection as our chosen solution.

The integration of a weakly supervised CNN character embedding noticeably enhanced accuracies of our DNNs. Still, we opted for retraining a FastText embedding on transactional text due to its enhanced computational efficiency and improved performance.

In sum up, our iterative experiments and ablation studies confirmed the TimeSeries+FastText approach results in an efficient and robust model architecture for transaction classification. The account offered in this appendix testifies to the validity of this chosen architecture over alternatives options.

\section{Expanded Results Discussion}
\label{app:recall}
\
We describe in this section the obtained recall metric for the chosen models described in \ref{sec:results}. 
% Increase row height by 10%
\renewcommand{\arraystretch}{1.2}
% Increase column separation by 10%
\setlength{\tabcolsep}{1.2\tabcolsep}

\begin{table}[ht!]
\centering
\scalebox{1.00}{
\begin{tabular}{|c|c|c|c|c|}
\hline
%\textbf{Models} & \textbf{Plaid API (Baseline)} & \textbf{Label Model} & \textbf{DNN Model} & \textbf{Improvement over Plaid} \\
\multirow{2}{*}{\textbf{Models}} & \multirow{2}{*}{\textbf{Plaid API}} & \textbf{Label} & \textbf{DNN} & \textbf{Improvement} \\
 & & \textbf{Model} & \textbf{Model} &  \textbf{Over Plaid} \\
\hline
Ads and Marketing & 0.32 & 0.44 & \underline{\textbf{0.93}} & +0.61 \\
\hline
Insufficient Funds Fees & 0.83 & 0.96 & \underline{\textbf{0.99}} & +0.16 \\
\hline
Wages and Payroll & 0.79 & \underline{\textbf{0.85}} & 0.79 & +0.06 \\
\hline
Personal & \textit{None} & \underline{\textbf{0.91}} & 0.90 & \textit{N/A} \\
\hline
Recurring & \textit{None} & 0.64 & \underline{\textbf{0.91}} & \textit{N/A} \\
\hline
Rent & 0.06 & 0.18 & \underline{\textbf{0.77}} & +0.71 \\
\hline
Revenue & \textit{None} & \underline{\textbf{0.98}} & 0.78 & \textit{N/A} \\
\hline
Telecoms & \underline{\textbf{0.96}} & 0.94 & 0.93 & -0.02 \\
\hline
Utilities & 0.41 & \underline{\textbf{0.98}} & 0.97 & +0.57 \\
\hline
\end{tabular}}
\caption{Recall for benchmark Plaid categories,
Label Model and DNN Model.}
\label{table:recall}
\end{table}

\renewcommand{\arraystretch}{1}
\setlength{\tabcolsep}{\tabcolsep}

Overall, each class best performing balanced accuracy model also corresponded to its best recall model. It is important to note that representative precision metrics could not be determined for our approach due to our mix of labeling sources (crowdsourcing and human manual annotation). These approaches change the test set distribution and class balances compared to the real world class balance and thus would give a false precision when assessed on these biased test sets. 

Ideally, the test sets should be generated by only randomly selecting samples for annotation and annotating them. However, with some class balances being close to 1\%, this would require to annotate a massive amount of  samples to obtain predictions for at least one hundred positive samples and a true estimation of the class balance. Moreover, not all samples can be annotated with certainty. These problems are the reason why we chose the balanced accuracy and recall metrics for our study, which, assuming that test set positive and negative examples are respectively randomly sampled from all positive and negative examples, is independent of the rebalancing of the test set. 

Seeing the results in tables \ref{table:results} and \ref{table:recall}, it is likely that Plaid's categorisation is tuned towards higher precision. Changing the decision threshold from 0.5, we show in table \ref{table:tune_precision} the recall metrics of our output scores with a decision threshold of 0.90. These results help compare our results to Plaid under the likely assumption that Plaid was tuned to achieve high precision.

% Increase row height by 10%
\renewcommand{\arraystretch}{1.2}
% Increase column separation by 10%
\setlength{\tabcolsep}{1.2\tabcolsep}

\begin{table}[ht!]
\centering
\scalebox{1.00}{
\begin{tabular}{|c|c|c|c|c|}
\hline
%\textbf{Models} & \textbf{Plaid API (Baseline)} & \textbf{Label Model} & \textbf{DNN Model} & \textbf{Improvement over Plaid} \\
\multirow{2}{*}{\textbf{Models}} & \multirow{2}{*}{\textbf{Plaid API}} & \textbf{Label} & \textbf{DNN} & \textbf{Improvement} \\
 & & \textbf{Model} & \textbf{Model} &  \textbf{Over Plaid} \\
\hline
Ads and Marketing & 0.32 & 0.49 & \underline{\textbf{0.88}} & +0.56 \\ %ok 
\hline
Insufficient Funds Fees & 0.83 & 0.92 & \underline{\textbf{0.99}} & +0.16 \\ %ok
\hline
Wages and Payroll & 0.79 & \underline{\textbf{0.90}} & 0.41 & +0.11 \\ %ok
\hline
Personal & \textit{None} & 0.49 & \textbf{\underline{0.75}} & \textit{N/A} \\ %ok
\hline
Recurring & \textit{None} & \textbf{\underline{0.47}} &  0.45 & \textit{N/A} \\ %ok
\hline
Rent & 0.06 & 0.18 & \underline{\textbf{0.72}} & +0.66 \\ %ok
\hline
Revenue & \textit{None} & \underline{\textbf{0.97}} & 0.54 & \textit{N/A} \\ %ok
\hline
Telecoms & 0.96 & \underline{\textbf{0.97}} & 0.91 & +0.01 \\ %ok
\hline
Utilities & 0.41 & \underline{\textbf{0.98}} & 0.84 & +0.57 \\  %ok
\hline
\end{tabular}}
\caption{Recall when increasing our decision thresholds to achieve high precision.}
\label{table:tune_precision}
\end{table}
\renewcommand{\arraystretch}{1}
\setlength{\tabcolsep}{\tabcolsep}

Interestingly,  the resulting findings differ from tables \ref{table:results} and \ref{table:recall}. When optimizing for precision, our models seem to show an incredibly large improvement (>55\%) over Plaid in terms of recall for the three categories where Plaid performs with less than 35\% recall. Plaid could possibly be overtuned towards precision for these three categories. Our Telecom model again seems to roughly match Plaid's. For the Insufficient Funds Fees and Wages categories, our models give a more modest +15\% and +11\% recall over Plaid. Also of note, for balanced accuracy, the Recurring DNN performed 9\% better than its Label Model, but when optimizing for precision, the Label Model performs 2\% better than the DNN. Inversely, the Personal DNN is better for precision optimisation (+26\% recall) compared to the Personal Label Model which is better for balanced accuracy (+4\% balanced accuracy).

These results seem to confirm our models can match or improve over Plaid's baselines.  Ideally, this claim should be confirmed with very large annotated sets of transactions, or by running in-the-field experimentation such as A/B tests (Plaid vs Model) for use cases and applications. However, these tests could not be accomplished in the scope of this paper.  Since our method also outputs continuous scores between 0 and 1, which is not the case in Plaid or other open banking data providers, our models can uniquely be tuned for specific precision-recall trade-offs. This precision-recall flexibility, improved balanced accuracy and large recalls for large decision thresholds lead us to believe that our models constitute an exciting and accurate option for bank transaction categorization, especially considering the cost-effectiveness of our methods.

\section{FastText Embedding}
\label{app:fasttext}

  \begin{figure}[ht!]
	\centering
	\includegraphics[scale=0.23]{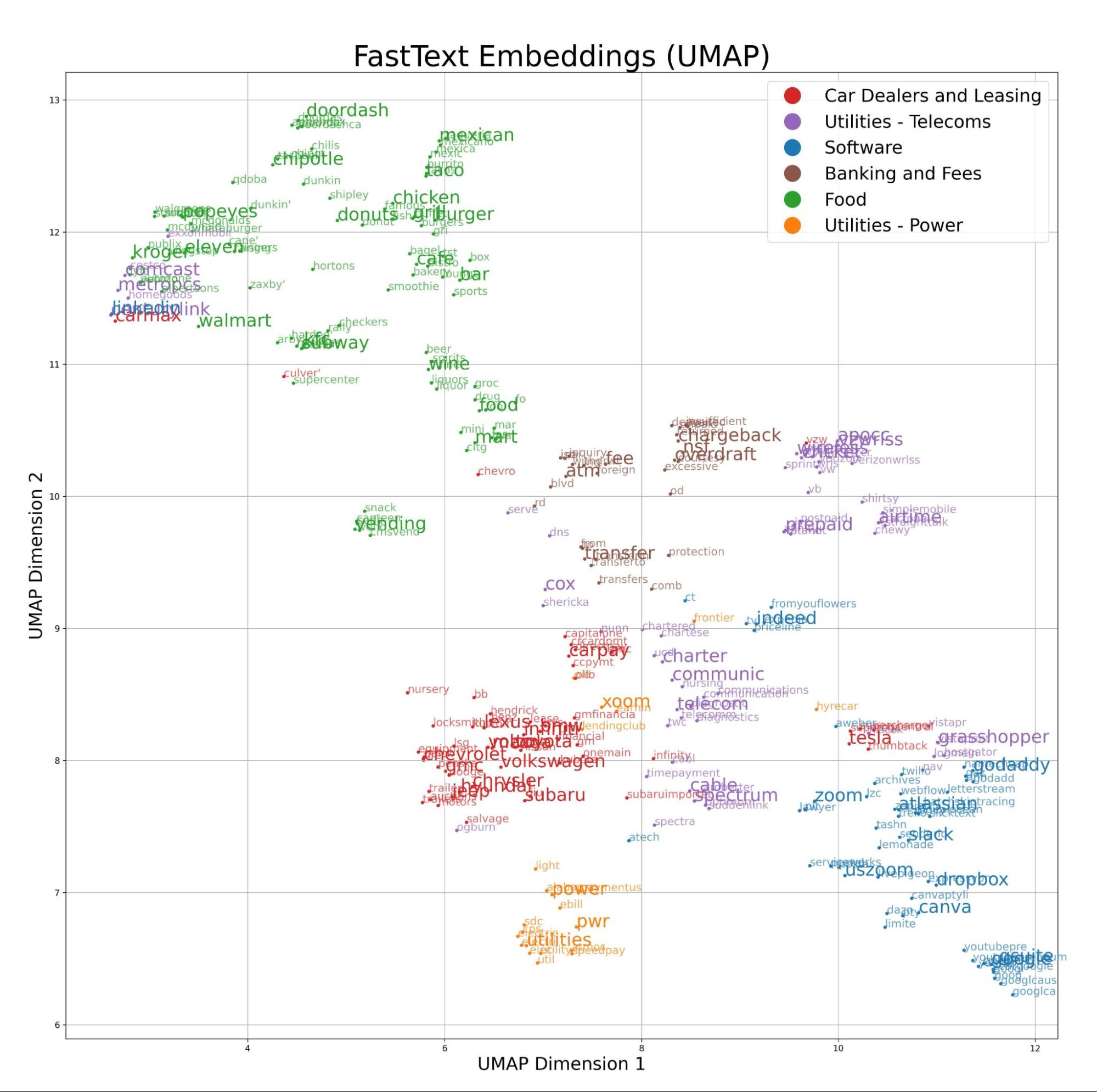}
	\caption{Transaction words and corresponding nearest
neighbors.}
	\label{fig:fasttext2}
\end{figure}

 We trained our FastText model using the popular python library gensim \cite{rehurek2011gensim}. Multiple FastText hyperparameters were experimented with, the most important being the underlying representation objective used to learn associations. Skip-gram representations outperformed continuous bags of words on our corpus.

The figure above is an enlarged version of the one presented in section \ref{sec:fasttext}. Fast Text word vectors for 6 categories were plotted in two dimensions using UMAP\cite{mcinnes2020umap}. Words with large fonts serve as the anchors, and smaller words around them are their nearest neighbors. To show the most prominent examples, we plot the 5 most-common nearest neighbours in a radius of 50 words to each anchor. These smaller words are still colored corresponding to the anchor’s category but are sometimes misclassified. Noticeably, anchors corresponding to the same category are often clustered together, and most neighbors do correspond to their anchor’s category, which makes anchoring a viable labeling function source. 

Other patterns are surprising, for instance, "tesla" is embedded closer to software companies which might indicate part of the transactions with the tesla keyword might be recurring transactions for subscriptions that unlock features in the cars. There is also a mixed category cluster on the top left, which corresponds to words for which the embedding did not create a meaningful word vector.

 Overall, training a new FastText custom embedding for transactions yielded a considerable improvement over the above-mentioned frozen or weakly-supervised, sparse, off-the-shelf, or character-level CNN concatenated embeddings. Moreover, FastText was very computationally efficient. Thanks to parallelization, it was possible to create word vectors for our total modeling corpus consisting of 18 million sentences which contain on average 44 characters, doing so in under 15 minutes.
\end{appendices}
\end{document}